\documentclass{article}

\usepackage{arxiv}

\usepackage[utf8]{inputenc} 
\usepackage[T1]{fontenc}    
\usepackage{hyperref}       
\usepackage{url}            
\usepackage{booktabs}       
\usepackage{amsfonts}       
\usepackage{amsmath}
\usepackage{nicefrac}       
\usepackage{microtype}      
\usepackage{cleveref}       
\usepackage{lipsum}         
\usepackage{graphicx}
\usepackage[numbers]{natbib}
\usepackage{doi}
\usepackage{parskip}

\usepackage{algorithm}
\usepackage{algpseudocode}

\usepackage[font=footnotesize]{caption}
\usepackage{subcaption}

\usepackage{geometry}
\usepackage{amsmath}
\usepackage{array} 
\usepackage{multirow}

\usepackage{threeparttable}

\DeclareMathOperator*{\argmax}{arg\,max}
\DeclareMathOperator*{\argmin}{arg\,min}

\title{Enhancing Performance and Calibration in Quantile Hyperparameter Optimization}

\date{}

\author{Riccardo Doyle \\ London, United Kingdom \\
 \texttt{r.doyle.edu@gmail.com}
}

\renewcommand{\headeright}{}
\renewcommand{\undertitle}{}

\hypersetup{
pdftitle={HPO Comparison},
pdfsubject={cs.LG},
pdfauthor={Riccardo Doyle},
pdfkeywords={hyperparameter optimization, HPO, benchmark, comparative, automl, tabular, tuning, conformal prediction, UCB, upper confidence bound sampling, bayesian optimization},
}

\begin{document}
\maketitle

\begin{abstract}
    Bayesian hyperparameter optimization relies heavily on Gaussian Process (GP) surrogates, due to robust distributional posteriors and strong performance on limited training samples. GPs however underperform in categorical hyperparameter environments or when assumptions of normality, heteroskedasticity and symmetry are excessively challenged. Conformalized quantile regression can address these estimation weaknesses, while still providing robust calibration guarantees. This study builds upon early work in this area by addressing feedback covariate shift in sequential acquisition and integrating a wider range of surrogate architectures and acquisition functions. Proposed algorithms are rigorously benchmarked against a range of state of the art hyperparameter optimization methods (GP, TPE and SMAC). Findings identify quantile surrogate architectures and acquisition functions yielding superior performance to the current quantile literature, while validating the beneficial impact of conformalization on calibration and search performance.
\end{abstract}


\section{Introduction} \label{Introduction}

Hyperparameter optimization algorithms seek to improve the convergence to a Machine Learning model's optimal hyperparameter configuration. Traditionally, in a single-fidelity setting, this is accomplished by training a surrogate model on accumulated configuration and performance pairs. The surrogate should display robust uncertainty quantification methods, allowing for acquisition functions capable of exploration and exploitation. Early examples \cite{NIPS2012_05311655} employ Gaussian Process (GP) \cite{Rasmussen2004} surrogates and Expected Improvement (EI) \cite{movckus1975bayesian} acquisition, outperforming random search \cite{JMLR:v13:bergstra12a}. 

GPs are particularly suited to hyperparameter optimization (HPO), due to high predictive performance on small datasets, extrapolation support, and a robust distributional framework. Several methods have however emerged to complement its weaknesses.

Tree Parzen Estimators (TPE) \cite{NIPS2011_86e8f7ab} provide a non-parametric, density-based alternative that improves GPs' natively poor handling of categorical features and unfavorable \(O(N^3)\) training time. While widely used, their performance in benchmarks is mixed and their non-parametric nature allows for only limited theoretical guarantees. 

SMAC \cite{10.1007/978-3-642-25566-3_40} utilizes a Random Forest \cite{breiman2014random} surrogate architecture, leveraging the mean and variance of individual tree predictions to parametrize a conditional posterior distribution for a configuration's performance. This is integrated with an Expected Improvement acquisition function. The use of a highly performant tree-based estimator is similarly aimed at improved handling of categorical features, however, its EI integration requires often unrealistic assumptions of normality and its parameterization is heuristic.

A more theoretically robust alternative to both TPE and SMAC can be found in quantile regression \cite{10.2307/1913643}. Work \cite{a05bb90f8fb8491392d9564461103e48, pmlr-v119-salinas20a} that explores quantile regression for hyperparameter optimization trains surrogates on pinball loss to obtain conditional quantile estimates per candidate configuration, which can inform probabilistic acquisition. While pinball loss provides calibration guarantees in the limit, it doesn't hold on finite horizons, resulting in at times limited applicability to the short horizons found in HPO.

To address the cumulative drawbacks of aforementioned GP competitors, recent work has focused on calibration of surrogate outputs via conformal prediction \cite{JMLR:v9:shafer08a}. In \cite{doyle2023achoadaptiveconformalhyperparameter}, a range of point and quantile estimators are calibrated via Locally Weighted Conformal Prediction \cite{Lei03072018} and Conformalized Quantile Regression (CQR) \cite{NEURIPS2019_5103c358} to provide finite sample guarantees. An optimistic upper quantile or upper prediction interval then guides acquisition. In \cite{salinas2023optimizing}, CQR calibrated Gradient Boosted Trees \cite{4a848dd1-54e3-3c3c-83c3-04977ded2e71} are used to generate evenly spaced quantile predictions, acting as a discrete conditional predictive distribution for each hyperparameter configuration, integrated with a Thompson Sampling \cite{dc35850b-2ca1-314f-9e0d-470713436b17} acquisition framework.

Compared to GPs, the surrogate flexibility of these approaches allows for better categorical feature handling and training time. Additionally, their quantile and conformal uncertainty framework allows for improved optimization performance in heteroskedastic or non-normal environments, while retaining distributional validity.

This paper seeks to deepen the success of conformalized quantile hyperparameter optimization frameworks, by addressing the following key gaps in current early work:
\begin{itemize}
    \item Existing benchmarks employ a broad pool of datasets, with limited regard for their attributes. In addition to evaluating frameworks on general benchmarks, this study performs dataset stratification to empirically validate whether quantile approaches outperform GPs on loss surfaces that most violate its assumptions of heteroskedastic errors and conditional symmetry.
    \item The range of explored acquisition functions in existing work is limited, with each work only exploring one function (optimistic Upper Confidence Bound Sampling \cite{journals/ml/AuerCF02} in \cite{doyle2023achoadaptiveconformalhyperparameter} and Thompson Sampling in \cite{salinas2023optimizing}). This paper evaluates the addition of Expected Improvement and Optimistic Bayesian Sampling \cite{JMLR:v13:may12a}.
    \item In \cite{salinas2023optimizing}, there is no attempt to control for covariate shift, resulting in potentially invalid conformal intervals and voiding of coverage guarantees. In \cite{doyle2023achoadaptiveconformalhyperparameter}, Adaptive Conformal Intervals (ACI) \cite{gibbs2021adaptive} are used to adjust conformal intervals in an online fashion, but analysis is limited to cumulative coverage on one sample dataset. This paper compares the use of ACI to the more robust Dynamically-tuned Adaptive Conformal Intervals (DtACI) \cite{JMLR:v25:22-1218}, as well as evaluating performance via a range of robust calibration metrics, across several datasets.
    \item A key benefit of conformal frameworks is surrogate model flexibility, however, existing work has not sufficiently explored the range of available surrogate architectures and their comparative performance. In \cite{salinas2023optimizing}, benchmarks are limited to a single architecture (Gradient Boosted Trees). In \cite{doyle2023achoadaptiveconformalhyperparameter}, benchmarks include a wider selection of architectures, but acquisition is limited to heuristic UCB Sampling and performances are only compared to random search on a limited range of datasets. This paper introduces previously unexplored architectures, such as post-hoc Quantile Gaussian Processes, Quantile Lasso \cite{51791361-8fe2-38d5-959f-ae8d048b490d} and quantile ensembles, as well as benchmarking all architectures proposed in \cite{doyle2023achoadaptiveconformalhyperparameter} against more competitive baselines and a wider range of datasets.
\end{itemize}

\section{Related Work}\label{RW}

\subsection{Conformalized Quantile Regression} \label{CQR}
Let us define some general training and validation sets as:
\begin{equation}
X_{train} ,Y_{train} =\{(X_i,Y_i) \mid {i\in \mathcal{I}_{train}, \mathcal{I}_{train} \not\subset \mathcal{I}_{cal}}\}
\end {equation}
\begin{equation}
X_{cal} ,Y_{cal} =\{(X_i,Y_i) \mid {i\in \mathcal{I}_{cal}, \mathcal{I}_{cal} \not\subset \mathcal{I}_{train}}\}
\end {equation}
Quantile regression \citep{10.2307/1913643} for some target quantile \(\beta\) involves training some conditional quantile estimator \(\hat Q_{\beta}(X)\) on \(X_{train}, Y_{train}\) via pinball loss \(L_\beta(u_i)\):
\begin{equation}
    L_\beta(u_i) =
    \begin{cases}
    u_i \beta & \text{if \(u_i > 0\)} \\
    u_i(\beta-1) & \text{if \(u_i \leq 0\)}
    \end{cases}
\end{equation}
Where \(u_i\) is the absolute error between \(\hat Q_{\beta}(X_i)\) and its target \(Y_i\) value.

An interval for a given \(X\) observation at a \(1-\alpha\) coverage can then be generated according to:
\begin{equation}
I(X) = [\hat Q_{\alpha/2}(X), \hat Q_{1-(\alpha/2)}(X)]
\end{equation}
While pinball loss provides limit guarantees, it does not ensure valid coverage on the finite \(X_{cal} ,Y_{cal}\) set. To remedy this, we can leverage Conformalized Quantile Regression \citep{NEURIPS2019_5103c358}. A set non-conformity scores can be generated based on validation set interval miss-coverage according to:
\begin{equation}
D_{cal} = \{max(\hat Q_{\alpha/2}(X_{i})-Y_{i}, Y_{i}- \hat Q_{1-(\alpha/2)}(X_{i})) \mid i \in \mathcal{I}_{cal}\}
\end{equation}
A new, calibrated \(1-\alpha\) interval can then be obtained by adjusting the original interval by the \(1-\alpha\) quantile of the non-conformity scores:
\begin{equation}
I(X) = [\hat Q_{\alpha/2}(X) - q_{1-\alpha}(D_{cal}), \hat Q_{1-(\alpha/2)}(X) + q_{1-\alpha}(D_{cal})]
\end{equation}

\subsection{Applications to Hyperparameter Optimization}\label{AHO}
Let us define some set of \(n\) randomly searched hyperparameter configurations and target model performances \(\{(X_t, Y_t)\}_{t=1}^n\).
We further split this into training and validation sets:
\begin{equation}
    X_{train} ,Y_{train} =\{(X_t,Y_t) \mid {t\in \mathcal{T}_{train}, \mathcal{T}_{train} \not\subset \mathcal{T}_{cal}}\}
\end{equation}
\begin{equation}
    X_{cal} ,Y_{cal} =\{(X_t,Y_t) \mid {t\in \mathcal{T}_{cal}, \mathcal{T}_{cal} \not\subset \mathcal{T}_{train}}\}
\end{equation}

Conformalized quantile regression surrogate frameworks then involve fitting some upper and lower quantile estimators \(\hat Q_{\alpha/2}(X)\) and \(\hat Q_{1-(\alpha/2)}(X)\) on  \(X_{train} ,Y_{train}\), and generating miss-coverage non-conformity scores on the validation set, as outlined in section \ref{CQR}:
\begin{equation} \label{quantile_non_conformity_scores}
D_{cal} = \{ max(\hat Q_{\alpha/2}(X_{t})-Y_{t}, Y_{t}-\hat Q_{1-(\alpha/2)}(X_{t})) \mid t \in \mathcal{T}_{cal}\}
\end{equation}
Which can then be used to generate a calibrated interval for \(X\) candidates:
\begin{equation} \label{quantile_conformal_interval}
I(X) = [\hat Q_{\alpha/2}(X) - q_{1-\alpha}(D_{cal}), \hat Q_{1-(\alpha/2)}(X) + q_{1-\alpha}(D_{cal})]
\end{equation}

Subsequent steps differ by framework. \citep{doyle2023achoadaptiveconformalhyperparameter} proposed sampling of the next hyperparameter configuration \(X_{n+1}\) via optimistic upper confidence bound sampling \cite{journals/ml/AuerCF02} of the interval in Eq. \ref{quantile_conformal_interval}: 
\begin{equation}\label{pessimistic_lower_bound}
X_{n+1}= \argmax_X (\{\hat Q_{1-(\alpha/2)}(X) + q_{1-\alpha}(D_{cal}) \mid X \in C\})
\end{equation}
Where \(C\) is the set of all unsampled hyperparameter configurations at \(n+1\). 

\citep{salinas2023optimizing} proposed sampling of the next hyperparameter configuration \(X_{n+1}\) via Thompson Sampling \cite{dc35850b-2ca1-314f-9e0d-470713436b17}. An even number of \(M\) equally spaced quantile estimators \(\{\hat Q_{\alpha_{i}}(X)\}_{i=1}^M\) are trained. For each symmetrical pair \(\{[\hat Q_{\alpha_{i}}(X), \hat Q_{\alpha_{M-i+1}}(X)]\}_{i=1}^{\frac{M}{2}}\), non-conformity scores (Eq. \ref{quantile_non_conformity_scores}) and calibrated intervals (Eq. \ref{quantile_conformal_interval}) are constructed. At \(n+1\), for each \(X\) in \(C\), a \(j \sim \mathcal{U}\{1, M\}\) is sampled, generating a calibrated quantile estimate of performance for each unsampled configuration of:
\begin{equation} \label{Eq14}
\hat Y(X) =
\begin{cases} 
\hat Q_{\alpha_{j}/2}(X) - q_{1-\alpha_{j}}(D_{cal}), & \text{if } j \leq \frac{M}{2} \\
\hat Q_{1-(\alpha_{j}/2)}(X) + q_{1-\alpha_{j}}(D_{cal}), & \text{otherwise}
\end{cases}
\end{equation}
Given their equal spacing, sampling uniformly from the quantile indices, then retrieving the quantile for that index, is equivalent to discretized inverse CDF sampling.

The next configuration to sample \(X_{n+1}\) is then determined by:

\begin{equation}
X_{n+1}= \argmax_X (\{\hat Y(X) \mid X \in C\})
\end{equation}

\section{Acquisition Extensions} \label{Samplers}
This section outlines a number of acquisition functions that have not yet been applied to conformal hyperparameter optimization.

\subsection{Expected Improvement (EI)}
Expected Improvement Sampling \cite{movckus1975bayesian} involves selecting a next best hyperparameter configuration \(X_{n+1}\) according to:
\begin{equation} \label{EI equation}
X_{n+1} = \argmax_X (\mathbb{E}\left[ \max(f(X) - f^*, 0) \right])
\end{equation}
Where \(f\) is some learned surrogate function, and \(\mathbb{E}\left[\max(f(X) - f^*, 0) \right]\) is the positively capped expectation of performance increase at some candidate configuration \(X\) over the previously achieved maximal performance \(f^*\).
Under a quantile search setting, the distribution of performance values at \(X\) can be discretely approximated by Monte Carlo sampling \(N\) observations from an even number of equally spaced quantile estimators \(\{\hat Q_{\alpha_{i}}(X)\}_{i=1}^M\), conformalized as seen in Eq. \ref{Eq14}. For each \(X\), the expected improvement is then the mean of the capped improvements across the \(N\) samples. Alternatively, given the deterministic nature of EI, the continuous calculation in equation \ref{EI equation} can simply be discretized over adjacent quantile intervals, assuming uniform density within intervals. For perfect quantile calibration, either approach will tend to the true Expected Improvement as \(M \to \infty\).

\subsection{Optimistic Bayesian Sampling}

Optimistic Bayesian Sampling (OBS) \cite{JMLR:v13:may12a} is an expectation floored variant of Thompson Sampling with favourable regret guarantees and empirical results. Assumptions underlying theoretical regret bounds don't hold in a Bayesian Optimization context, so it's provided as a heuristic modification.

To outline methodology, let \(P(Y|X_{i})\) represent some posterior distribution for performance \(Y\) under configuration \(X_{i}\). Further let \(Y_{X_{i}} \sim P(Y|X_{i})\) represent a randomly sampled realization from the posterior. OBS promotes exploratory behaviour by bounding the realization by the conditional expectation \(\hat f(X_{i}) \rightarrow \mathop{\mathbb{E}}(Y | X)\), resulting in a final realization of \(\max(\hat f(X_{i}), Y_{X_{i}})\). This forces the initial Thompson Sampling realizations toward positive uncertainty regions of the posterior, rewarding high variance configurations. Exploratory pressure can help in short, finite horizons, though only empirical benchmarks can inform whether greater exploration is beneficial or detrimental.

\section{Surrogate Extensions}\label{Surrogate Extensions}
\subsection{Surrogate Architectures} 
In addition to the quantile Gradient Boosted Machine (QGBM) \cite{4a848dd1-54e3-3c3c-83c3-04977ded2e71} surrogate architecture seen in \cite{salinas2023optimizing} and \cite{doyle2023achoadaptiveconformalhyperparameter}, and the Quantile Regression Forest (QRF) \cite{JMLR:v7:meinshausen06a} seen in \cite{doyle2023achoadaptiveconformalhyperparameter}, the following architectures are explored:
\begin{itemize}
    \item Quantile Lasso (QL): A simple Lasso estimator \cite{51791361-8fe2-38d5-959f-ae8d048b490d} trained via pinball loss, with the potential to better handle high dimensionality, linear loss surfaces and to avoid excessive overfitting in early search.
    \item Quantile Gaussian Process (QGP): A Gaussian Process \cite{Rasmussen2004} estimator whose posterior is used to extract empirical quantiles, for compatibility with Equation \ref{quantile_conformal_interval}. This allows comparison of Gaussian Processes against alternative surrogate architectures under the same conformal framework, although discretization of the posterior into quantiles may degrade performance.
\end{itemize}

\subsection{Ensembling} 
Ensembles of above architectures are introduced to reduce overfitting on small surrogate training datasets and increase generalization across different hyperparameter loss surfaces.

An ensemble of \(M\) base quantile estimators \(Q^1_\beta,\dots,Q^M_\beta\) targeting some quantile \(\beta\), can be obtained via quantile linear stacking \cite{WOLPERT1992241}. First, each observation \((X_i, Y_i)\) is assigned to some corresponding cross validation fold \(S_{k(i)}\). For a given base estimator \(Q^m_\beta\), a hold out fold prediction can be obtained as:
\begin{equation}
\qquad z_{i,m}=Q^m_{\beta, -S_{k(i)}}(x_i)
\end{equation}
Where \(Q^m_{\beta, -S_{k(i)}}\) denotes an estimator not trained on fold \(-S_{k(i)}\). The predicted quantiles of each base estimator then form the new features of a Quantile Lasso meta learner, selecting a weight vector that minimizes the pinball loss \(L_\beta\) between stacked predictions and original \(Y\) targets:
\begin{equation}
\argmin_{w_m}
\frac{1}{n}\sum_{i=1}^n
L_\beta\!\Bigl(y_i-\sum_{m=1}^M x_{i,m}w_m\Bigr)
+\lambda\sum_{m=1}^M|w_m|
\quad\text{s.t.}\quad w_m\ge0\;\;\forall m.
\end{equation}

Weights are positively constrained to reduce instability on small datasets.

Based on complementary strengths and avoidance of excessive multiple comparisons, we propose a single versatile ensemble architecture (though more could be systematically explored):
\begin{itemize}
    \item \textit{QE}: An ensemble of QGBM, QL and QGP architectures. QGBM provides strong tree-based categorical feature handling; QL provides dimensionality reduction and support for linear relationships; while QGP provides high performance in low observation environments and powerful distributional priors (when corectly specified).
\end{itemize}

\section{Conformal Extensions}

\subsection{Sample Efficiency}\label{Sample Efficiency}
Existing hyperparameter optimization applications outlined in section \ref{CQR} partition all available data into training and calibration sets. This trades training quality for calibration quality, which may result in loss of important training patterns in a low observation HPO context. Additionally, regardless of proposed split trade off, calibration sets are generally limited in size, particularly in early search, and have the potential to bias adjustments. 
CV+ \cite{barber2020predictiveinferencejackknife} can be utilized to lessen or eliminate the loss of training information, while retaining calibration guarantees. Implementation involves splitting available data into \(K\) folds \(S_1, \ldots, S_K\), and obtaining fold-specific non-conformity scores as:
\begin{equation}
D_i = \{ max(\hat Q^{-S_{k(i)}}_{\alpha/2}(X_{i})-Y_{i}, Y_{i}-\hat Q^{-S_{k(i)}}_{1-(\alpha/2)}(X_{i})) \mid i \in \{1, 2, \ldots, n\}
\end{equation}
Where \(Q^{-S_{k(i)}}\) represents a quantile regression trained on all folds except \(S_{k(i)}\), and \(k(i)\) represents the fold containing the observation indexed by \(i\).

A prediction interval for some next sampled configuration \(X_j\) can then be generated by making a prediction for the configuration using each fold's model, adjusting that prediction by each non-conformity score in the model's holdout fold, then taking the \(1-\alpha\) quantile of adjusted predictions:

\begin{equation}
I(X_j) = [q_{1-\alpha}(\hat Q^{-S_{k(j)}}_{\alpha/2}(X_j) - D_{j}), q_{1-\alpha}(\hat Q^{-S_{k(j)}}_{1-(\alpha/2)}(X_j) + D_{j})]
\end{equation}

For \(K\) approaching \(n\), this approach minimizes loss of training data while still providing valid coverage guarantees. It is however impractical to set such a large \(K\) due to surrogate re-training costs. In this study we set \(K=5\) when using CV+, and separately introduce a heuristic adaptive method that leverages CV+ in early search (\(t<50\)), switching to split conformal prediction thereafter.

\subsection{Addressing Feedback Covariate Shift} \label{FCS}

Conformal prediction requires exchangeability of non-conformity scores, which, however, is not guaranteed in a sequential hyperparameter optimization setting. To ensure conformal intervals generated by surrogate models remain valid, Adaptive Conformal Intervals (ACI) \citep{gibbs2021adaptive} can be generated by adjusting the miss-coverage level \(\alpha\) after each sampled configuration according to:
\begin{equation} \label{eq aci}
\alpha_{t+1} = \alpha_t + \gamma(\alpha - \epsilon_t)
\end{equation}
Where \(\gamma\) is a tunable learning rate and \(\epsilon_t\) is a binary miss-coverage indicator. In the literature, this adjustment is applied in \cite{doyle2023achoadaptiveconformalhyperparameter}, but not in \cite{salinas2023optimizing}.

This study correctly applies ACI, while also comparing it to the theoretically more robust Dynamic Tunable Adaptive Conformal Interval (DtACI) \cite{JMLR:v25:22-1218} framework, which replaces the fixed \(\gamma\) parameter in Eq. \ref{eq aci} with a set of \(K\) candidate values \(\{\gamma_i\}_{i=1}^K\) and corresponding candidate miscoverage levels \(\{\alpha_i\}_{i=1}^K\). 

The implementation of DtACI involves, at \(t=1\), the initialization of a unit vector \(w_{t=1}^{i} = 1\) and starter misscoverages \(\alpha_{t=1}^{i} = \alpha\) for each candidate, as well as a starter consensus misscoverage \(\alpha_t = \alpha\). At \(t=2\) an observation is sampled and empirical interval feedback is obtained as:
\begin{equation}
\beta_t := \sup \left\{ \beta : Y_t \in \hat{C}_t(\beta) \right\}
\end{equation}

Where \(\hat{C}_t(\beta)\) is the smallest \(\beta\) confidence interval containing the sampled observation \(Y_t\). Weights are then updated based on the pinball loss between \(\beta\) and each candidate misscoverage \(\alpha_t^i\):
\begin{equation}
\bar{w}^i_t \gets w^i_t \exp(-\eta \ell(\beta_t,\alpha_t^i))
\end{equation}
Where \(\eta\) is a tunable parameter. Weights are further regularized as:
\begin{equation}
w^i_{t+1} \gets (1-\sigma) \bar{w}^i_t + \frac{(\sum_{1 \leq i \leq K} \bar{w}^i_t) \sigma}{K}
\end{equation}

Where \(\sigma\) is a tunable parameter.

Actual misscoverage indicators \(\epsilon^i_t\) between the sampled point \(Y_t\) and every candidate interval \(\hat{C}_t(\alpha^i_t)\) are then obtained, resulting in the updated candidate misscoverage levels:
\begin{equation}
\alpha_t = \alpha^i_{t} + \gamma_i(\alpha - \text{err}^i_{t})
\end{equation}

From which the next shift adjusted \(\alpha_t\) is sampled proportionately to a distribution defined by the normalized candidate weights \(\frac{w^i_t}{\sum_{1 \leq j \leq K}w^{j}_t}\).

In alignment with the original paper, this study sets \(\sigma\) and \(\eta\) parameters to:
\begin{equation}
\eta = \sqrt{\frac{3}{L} \cdot \frac{\log(L K) + 2}{(1-\alpha)^2 \alpha^2}} \qquad \sigma = 1/(2L)
\end{equation}

Where \(L\) is the local interval length, with higher \(L\) resulting in a tighter regret bound, but possibly weaker local coverage, and vice versa. This study sets \(L\) arbitrarily to 50 when considering an experiment horizon of 100 trials.

\section{Benchmarking} \label{Benchmarking}
\subsection{Environments}
The performance of previously outlined enhancements will be assessed across three core benchmarking environments:
\begin{itemize} 
    \item \textbf{JAHS-Bench-201} \cite{NEURIPS2022_fd78f2f6}: Neural Network architecture optimization spanning 2 continuous and 9 categorical hyperparameters across \textit{CIFAR-10} \cite{krizhevsky2009learning}, \textit{Colorectal-Histology} \cite{kather2016multi}, and \textit{Fashion-MNIST} \cite{Xiao2017FashionMNISTAN} image recognition datasets. 
    \item \textbf{LCBench} \cite{ZimLin2021a}: Neural Network architecture optimization spanning 4 continuous and 3 integer hyperparameters across 35 tabular \textit{OpenML} datasets.
    \item \textbf{rbv2\_aknn} \cite{pmlr-v188-pfisterer22a}: Hyperparameter optimization spanning 4 integer and 2 categorical hyperparameter across 119 tabular \textit{OpenML} datasets, relating to an Approximate Nearest Neighbours \cite{10.1109/TPAMI.2018.2889473} classification task.
\end{itemize}
Given their size, \textit{LCBench} and \textit{rbv2\_aknn} are not benchmarked in full. Rather, their extensive dataset count is leveraged to create experimental ML sub-populations displaying three characteristics of interest:
\begin{itemize} 
    \item \textbf{Size}: For each dataset, we sample 10,000 hyperparameter configurations, average the runtime required to train on each configuration, and select the 5 datasets with the highest average runtime. This selection focuses on expensive, slow datasets on which hyperparameter optimization is most likely to be applied in practice. Size benchmarks are referenced as \textbf{LCBench-L} and \textbf{rbv2\_aknn-L}.
    \item \textbf{Residual Heteroskedasticity}: For each dataset, we sample 10,000 hyperparameter configurations and performances, fit a Gaussian Process and obtain point-estimate residuals. An auxiliary linear regression predicts squared residuals using configurations to quantify heteroskedasticity. The 5 most heteroskedastic datasets by adjusted \(R^2\) are included in this sub-population. This selection focuses on loss surfaces with the highest breaches of traditional GP assumptions, to gauge added benefit of non-distributional quantile regression surrogates. Heteroskedasticity benchmarks are referenced as \textbf{LCBench-H} and \textbf{rbv2\_aknn-H}.
    \item \textbf{Conditional Asymmetry}: For each dataset, we sample 10,000 hyperparameter configurations and performances. A K-Nearest Neighbours (KNN) \cite{1053964} model estimates local performance spread at each configuration. The 5 most asymmetric datasets by average absolute quantile skew across all configurations are included in this sub-population. This selection focuses on loss surfaces that breach GP symmetry assumptions, but are well suited to quantile regression's independent quantile estimation. Asymmetry benchmarks are referenced as \textbf{LCBench-A} and \textbf{rbv2\_aknn-A}.
\end{itemize}

All above benchmarking environments are accessed via tree based surrogate estimators provided by \textit{jahs\_bench\_201} \cite{NEURIPS2022_fd78f2f6} and \textit{yahpo\_gym} \cite{pmlr-v188-pfisterer22a} Python packages.
\textit{OpenML} identifiers of sub-population benchmarking environments can be found in Appendix \ref{Appendix - OpenML Stratifications}.

All benchmarking environments are evaluated at full fidelity.

\subsection{Metrics}
Hyperparameter optimization frameworks will be assessed on the basis of the aforementioned benchmarking environments, utilizing the goal metric of each dataset in the benchmark (most often validation accuracy of the sampled configuration).

For each dataset, a given framework is rerun \(n\) times, with \(n\) being specified in the results section per type of analysis. For each run, a value for cumulative best performance is computed at each iteration. Ranks for a given run are calculated at each iteration, based on the relative performance of other frameworks at that iteration. Lastly, performances and ranks are averaged (or otherwise aggregated) across runs, by iteration, resulting in a single optimization result path per framework, per dataset. These results may be further aggregated at benchmark level, depending on the analysis type. 

This sequence of operations can be applied at either iteration or runtime budget level, with this study focusing on iteration level (given the lack of explicit multi-objective runtime optimization), but runtime aggregations for each simulation are provided in Appendix \ref{Appendix - Runtime Aggregated Benchmarks}.

\subsection{Parameters}
All surrogate models and acquisition functions have default parameters, details of which can be found in the source code (Appendix \ref{Appendix - Code}). However, below is a list of key benchmark-specific parameters worth noting:
\begin{itemize} 
    \item \textbf{Random Trials}: All HPO algorithms require an initial number of randomly sampled configuration and performance pairs to train on; this number was set to 15 for all benchmarks. For a given repetition, all models will receive the same 15 warm starts.
    \item \textbf{Budget}: All HPO algorithms are run for a total of 100 iterations, with search performance later reported at both iteration and runtime budget levels.
    \item \textbf{Candidate Space}: All HPO algorithms pass all candidate configurations from the search space to their acquisition function to make a sampling decision. This is expensive, or intractable for large search spaces, so a random sample of size \(n\) is taken from the search space instead. For all benchmarks and algorithms, \(n=2000\).
    \item \textbf{Minimum Observation Count for Conformalization}: Conformalized quantile surrogates begin to train and infer on the first 15 warm starts, however the process of conformalizing surrogate predictions does not begin until a later total number of configurations are sampled (as seen in \cite{salinas2023optimizing}). This number is set to 32, to avoid observation loss and miss-calibration when data availability is low.
\end{itemize}

Additionally, all conformalized quantile algorithms are trained using SCP and DtACI unless otherwise stated. The use of SCP over CV+ eases runtime burden.

\section{Results} \label{Results}

\subsection{Calibration} \label{calibration results}

Table \ref{tab:calibration_metrics_by_entity} compares unconformalized, CV+ and SCP quantile regression, with and without adaptive adjustment, across a range of calibration metrics.

All aforementioned variants derive their samples from 15 random draws followed by expectation maximization of a QGBM surrogate. The deterministic sampling ensures no search effect contamination on coverage and creates immediate distributional shift between random and greedy phases.

Though conformal prediction only provides marginal guarantees, results suggest strong local calibration benefit, with all conformal variants achieving lower ranks than the unconformalized variant. Among them, CV+ outperforms SCP, while, regardless of conformalization framework, adaptation benefits conformalization, with DtACI outperforming ACI.

In addition to temporal local calibration quality, Log-Likelihood Ratios (LLRs) are reported to provide a view of feature space conditioning. In this regard, effect size is more contained, with smaller rank spreads than in previous results. Outcomes are also more mixed, with SCP improving base conditional calibration, and CV+ worsening it. Additionally, DtACI provides consistent performance improvements, but ACI does not. It is worth noting that Table \ref{tab:calibration_metrics_by_entity} compares ranks, not magnitudes, so a given comparison may contain a significant rank difference, even if all raw LLR statistics in the comparison are insignificant (meaning both compared variants might exhibit no or low nominal correlation between breaches and features).

\begin{table}[htbp]
\centering
\begin{threeparttable}
\captionsetup{skip=10pt}
\caption{Calibration performance rank by calibration metric. Metrics are computed for intervals at 25\%, 50\% and 75\% confidence on all \textit{LCbench} datasets, then ranked across frameworks within each interval confidence and dataset. Individual ranks are then averaged by framework to demonstrate cross-confidence and cross-dataset performance.}
\label{tab:calibration_metrics_by_entity}
\begin{tabular}{@{}l*{3}{>{\centering\arraybackslash}p{3.45cm}}@{}}
\toprule
\textbf{} & \textbf{Rolling Coverage Error Rank\tnote{1}} & \textbf{LLR Statistic Rank\tnote{2}} & \textbf{Interval Width Rank\tnote{3}} \\
\midrule
\normalsize{\textbf{Unconformalized}} & \begin{minipage}{3.45cm}\centering \normalsize{5.283} \\ \small{[5.113, 5.450]} \end{minipage} & \begin{minipage}{3.45cm}\centering \normalsize{3.893} \\ \small{[3.717, 4.117]} \end{minipage} & \begin{minipage}{3.45cm}\centering \normalsize{\textbf{1.307}} \\ \small{[1.213, 1.400]} \end{minipage} \\

\normalsize{\textbf{Split Conformalized}} & \begin{minipage}{3.45cm}\centering \normalsize{4.322} \\ \small{[4.137, 4.554]} \end{minipage} & \begin{minipage}{3.45cm}\centering \normalsize{3.650} \\ \small{[3.432, 3.838]} \end{minipage} & \begin{minipage}{3.45cm}\centering \normalsize{4.797} \\ \small{[4.477, 5.167]} \end{minipage} \\
\normalsize{\quad + ACI} & \begin{minipage}{3.45cm}\centering \normalsize{4.195} \\ \small{[4.048, 4.425]} \end{minipage} & \begin{minipage}{3.45cm}\centering \normalsize{3.703} \\ \small{[3.498, 3.903]} \end{minipage} & \begin{minipage}{3.45cm}\centering \normalsize{4.627} \\ \small{[4.460, 4.810]} \end{minipage} \\
\normalsize{\quad + DtACI} & \begin{minipage}{3.45cm}\centering \normalsize{3.600} \\ \small{[3.492, 3.732]} \end{minipage} & \begin{minipage}{3.45cm}\centering \normalsize{\textbf{3.567}} \\ \small{[3.370, 3.752]} \end{minipage} & \begin{minipage}{3.45cm}\centering \normalsize{4.563} \\ \small{[4.386, 4.760]} \end{minipage} \\

\normalsize{\textbf{Cross Conformalized}} & \begin{minipage}{3.45cm}\centering \normalsize{4.058} \\ \small{[3.967, 4.143]} \end{minipage} & \begin{minipage}{3.45cm}\centering \normalsize{4.440} \\ \small{[4.258, 4.628]} \end{minipage} & \begin{minipage}{3.45cm}\centering \normalsize{4.397} \\ \small{[4.237, 4.567]} \end{minipage} \\
\normalsize{\quad + ACI} & \begin{minipage}{3.45cm}\centering \normalsize{3.657} \\ \small{[3.557, 3.742]} \end{minipage} & \begin{minipage}{3.45cm}\centering \normalsize{4.487} \\ \small{[4.282, 4.703]} \end{minipage} & \begin{minipage}{3.45cm}\centering \normalsize{4.253} \\ \small{[4.033, 4.457]} \end{minipage} \\
\normalsize{\quad + DtACI} & \begin{minipage}{3.45cm}\centering \normalsize{\textbf{2.885}} \\ \small{[2.663, 3.080]} \end{minipage} & \begin{minipage}{3.45cm}\centering \normalsize{4.260} \\ \small{[4.158, 4.385]} \end{minipage} & \begin{minipage}{3.45cm}\centering \normalsize{4.057} \\ \small{[3.653, 4.407]} \end{minipage} \\
\bottomrule
\end{tabular}
\vspace{1em}
\begin{tablenotes}
\footnotesize
\item[1] Rank of average coverage error across non-overlapping windows of 20 consecutive search iterations. 
\item[2] Rank of log-likelihood ratio from Logistic Regression training on hyperparameter values \(X\) and binary interval breach indicator \(Y\), for each sampled hyperparameter configuration and corresponding interval pair.
\item[3] Rank of average conformal prediction interval width, across sampled configuration intervals.
\end{tablenotes}
\end{threeparttable}
\end{table}

Turning to interval quality, interval widths are significantly larger in conformalized variants. Among them, CV+ reduces widths meaningfully compared to SCP, with adaptation reducing widths further, regardless of conformal framework.

For a more visual interpretation of findings in Table \ref{tab:calibration_metrics_by_entity}, Appendix \ref{Appendix - Calibration Performance by LCBench-L Dataset} provides a comprehensive breakdown of cumulative coverage error across \textit{LCBench-L} datasets, conformal variants and confidence levels. Interestingly, conformalization benefit is strongest for 50\% and 75\% confidence levels, with mixed to poor results at 25\% confidence. This can be attributed to the noisier behaviour of non-conformity scores as intervals become excessively narrow.

\subsection{Acquisition}

\paragraph{Acquisition Function Comparison}

\begin{figure}[htbp]
    \centering
    \includegraphics[scale=0.5]{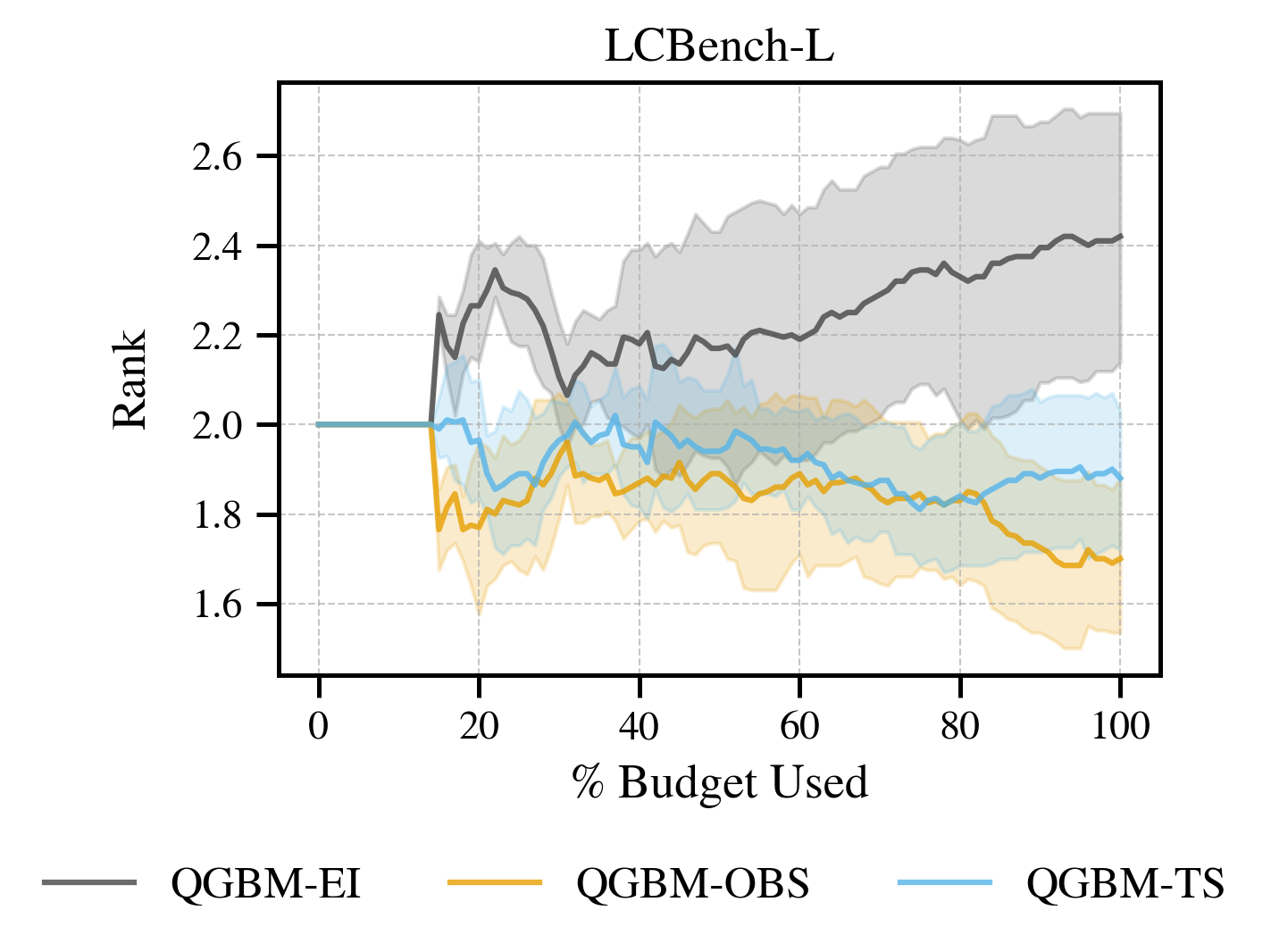}
    \caption{Search performance rank over iteration search budget per acquisition function on \textit{LCBench-L}, across 20 random warm start initializations. Shaded region represents 95\% dataset-bootstrapped interval.} 
    \label{af-iteration}
\end{figure}

Existing literature has provided only limited exploration of acquisition functions. To supplement it, this section takes the literature's best performing surrogate architecture (QGBM) and compares previously benchmarked Thompson Sampling (TS) acquisition to Expected Improvement (EI) and Optimistic Bayesian Sampling (OBS). Search performance across \textit{LCBench-L} datasets is provided in Figure \ref{af-iteration}. 

Findings suggest EI strongly underperforms Thompson alternatives. This could be explained by the quantile discretization poorly capturing tail end behaviour, which can be crucial for Expected Improvement acquisition, particularly as search progresses and the best historical performance continues to improve.

Between Thompson Sampling approaches, OBS outperforms standard TS, though the difference is not as marked as that between EI and Thompson approaches. This suggest there is repeatable benefit to greater exploration across the selected subset of \textit{LCBench-L} datasets.

\begin{figure}[htbp]
    \centering
    \includegraphics[scale=0.5]{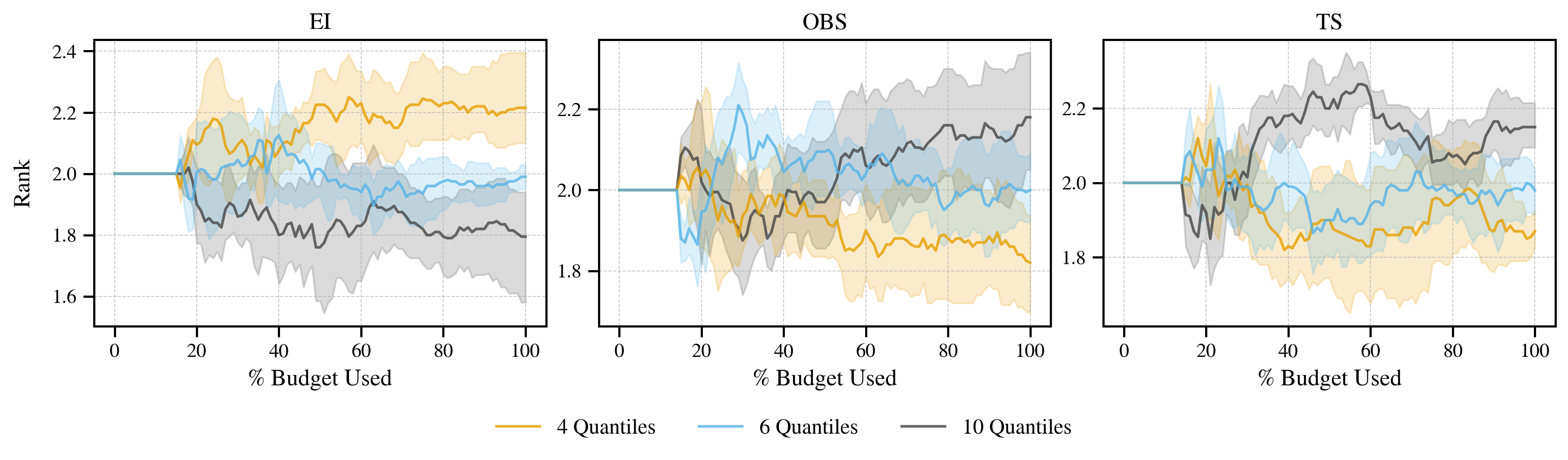}
    \caption{\textit{LCBench-L} search performance rank over iteration search budget for a QGBM surrogate, across multiple acquisition functions (columns) and quantile densities. Results cover 20 random warm start initializations. Shaded region represents 95\% dataset-bootstrapped interval.} 
    \label{nq-iteration}
\end{figure}

\paragraph{Quantile Density} Existing distributional quantile regression approaches \cite{salinas2023optimizing} utilize 4 quantiles to approximate the conditional distribution at \(X\). To briefly explore the impact of this choice, Figure \ref{nq-iteration} analyzes the change in search performance as the number of quantiles is increased from 4 to 10. 

The expectation is that a higher number of quantiles reduces the approximation error between the discretized quantile distribution and the true distribution. Interestingly, this is true under an Expected Improvement (EI) sampling regime, but not true of the Thompson Sampling approaches, where quantile count is independent of performance. This can be explained by EI's determinism and increased reliance on tail behaviour as the observed optimum improves with search time, resulting in improved performance as a larger number of quantiles starts to capture extreme distribution regions. Thompson Sampling instead involves heavy randomness and makes use of the entire distribution range throughout search, reducing dependance on distribution granularity, particularly if extreme quantiles or increased step density don't meaningfully alter uncertainty allocation.

\subsection{Surrogate Architecture}\label{Surrogate Architecture}

Analysis has so far focused on the literature's currently best performing surrogate (QGBM). This section provides an assessment of how different surrogates compare to each other, and whether any approaches improve on the current state of the art (SOTA).

\begin{figure}[htbp]
    \centering
    \includegraphics[scale=0.5]{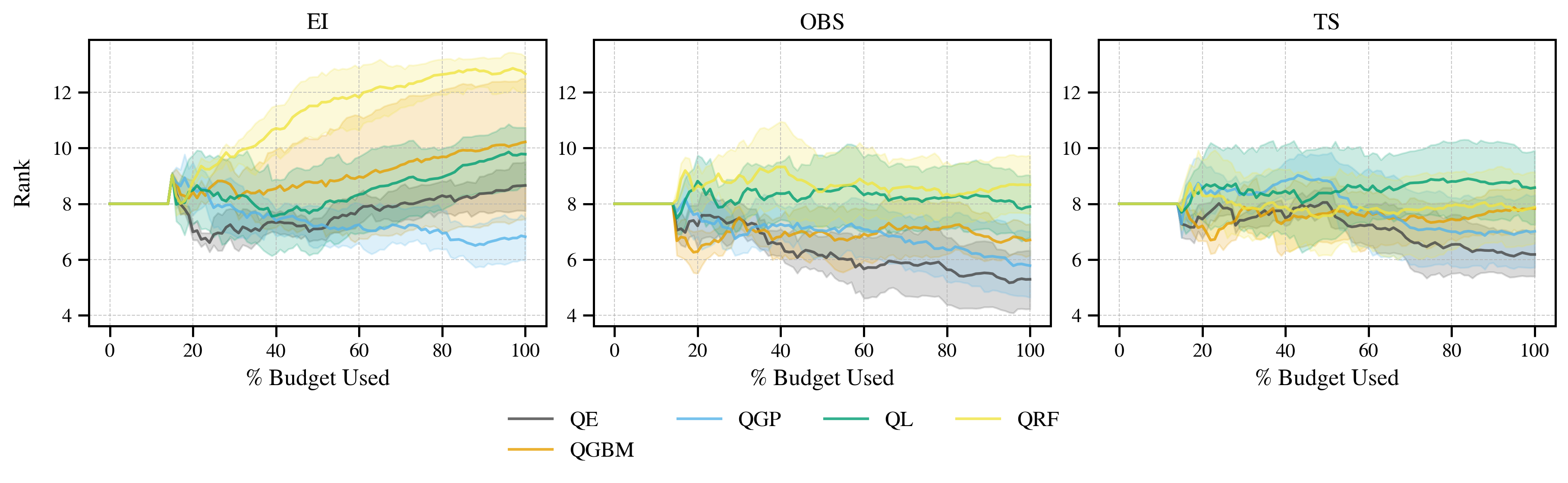}
    \caption{\textit{LCBench-L} search performance rank over iteration search budget for a range of surrogate architectures, across multiple acquisition functions (columns). Ranks are shared across plots (each surrogate and acquisition combination is treated as a ranking variant). Results cover 20 random warm start initializations. Shaded region represents 95\% dataset-bootstrapped interval.} 
    \label{dynamic-arch-iteration}
\end{figure}

Figure \ref{dynamic-arch-iteration} displays the search performance of surrogate architectures outlined in section \ref{Surrogate Extensions} across various acquisition functions on \textit{LCBench-L}. With the exception of QGP, surrogates generally underperform when sampling via Expected Improvement. QGP's resilience to EI may be due to greater quantile estimation accuracy, particularly at extreme quantiles. QE outperforms other architectures on OBS and TS, and results in the most consistent performance across acquisition functions, highlighting the benefits and versatility of ensembles.

Though it differs by acquisition function, QGP and QGBM display the strongest aggregate performance outside of QE, with QRF and QL frequently competing for last position.

\begin{figure}[htbp]
    \centering
    \includegraphics[scale=0.5]{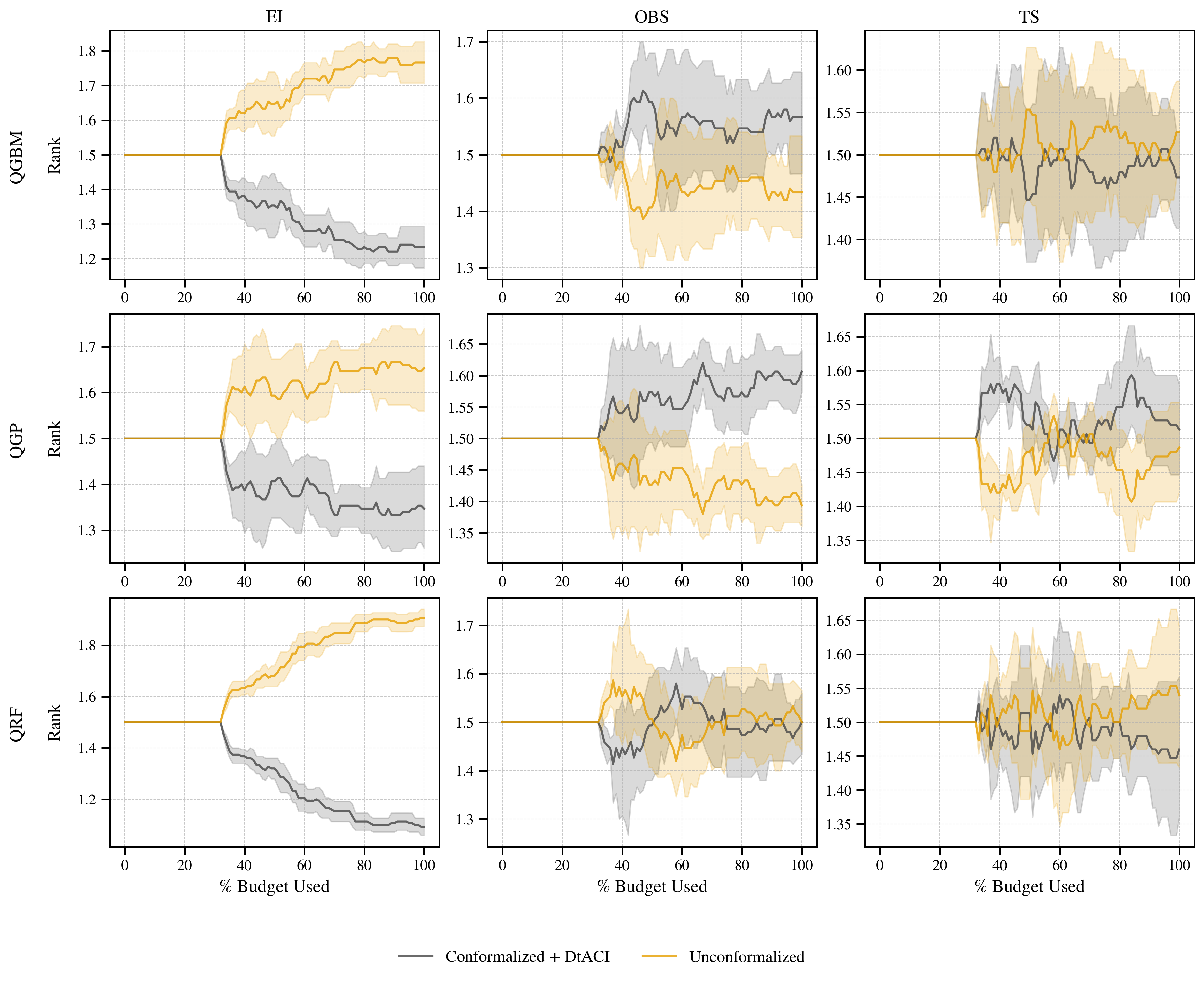}
    \caption{\textit{LCBench-L} search performance rank over iteration search budget. Performances are reported with and without conformalization, across several surrogate architectures (rows) and acquisition functions (columns). Results cover 20 random warm start initializations. Shaded region represents 95\% dataset-bootstrapped interval. Conformalization is carried out via CV+ up to the 50-th iteration, and SCP thereafter.} 
    \label{ce_iteration}
\end{figure}

\paragraph{Conformalization Impact} Section \ref{calibration results} demonstrated the calibration benefits of conformalization, however this doesn't necessarily translate to empirical search performance. To quantify whether conformalization results in more robust search, Figure \ref{ce_iteration} compares a range of surrogate architectures trained with and without conformalization, across a range of acquisition functions on \textit{LCBench-L}. 

Findings indicate strong distinctions between EI and Thompson approaches, with the former displaying both large and significant benefits from conformalization and the latter hovering between insignificance and negative impact. There is limited heterogeneity of effect between surrogate architectures, with the exception of QGP, which displays noticeably smaller EI conformalization benefits than QGBM and QRF (while remaining beneeficial and significant). This may be due to some datasets adhering well to GP assumptions, resulting in conformalization adding extra noise or bias compared to a more consistently beneficial effect in poorly calibrated tree estimators.

The insignificant, to occasionally negative, impact of conformalization under Thompson approaches can have several causes. Better calibration can hurt search performance if model misspecification suits the search environment. A surrogate that consistently underestimates quantiles, with clustering around the mean, will outperform a correctly calibrated one in an extremely greedy environment (or inversely, an overestimating surrogate in an exploratory environment). 

Conformalization may also be providing better marginal coverage, without improving, or, while worsening, conditional coverage, leading to worse search performance, though evidence of this is limited, given strong EI benefit and low local coverage error in earlier analysis.

Lastly, the loss of training data resulting from Split Conformal Prediction may be harming inference, though this is also not supported by evidence, since Figure \ref{ce_iteration}'s charts were generated via adaptive schedule conformalization, and no clear rank shift is detectable when switching from CV+ to SCP past iteration 50.

\pagebreak

\subsection{SOTA Comparison} 

\paragraph{General Analysis} Previous sections have identified several architectures and acquisition functions capable of outperforming current quantile conformal approaches. In this section, a subset of those architectures is benchmarked on a more exhaustive spread of datasets and compared to popular alternative HPO algorithms.

\begin{figure}[t]
    \centering
    \includegraphics[scale=0.5]{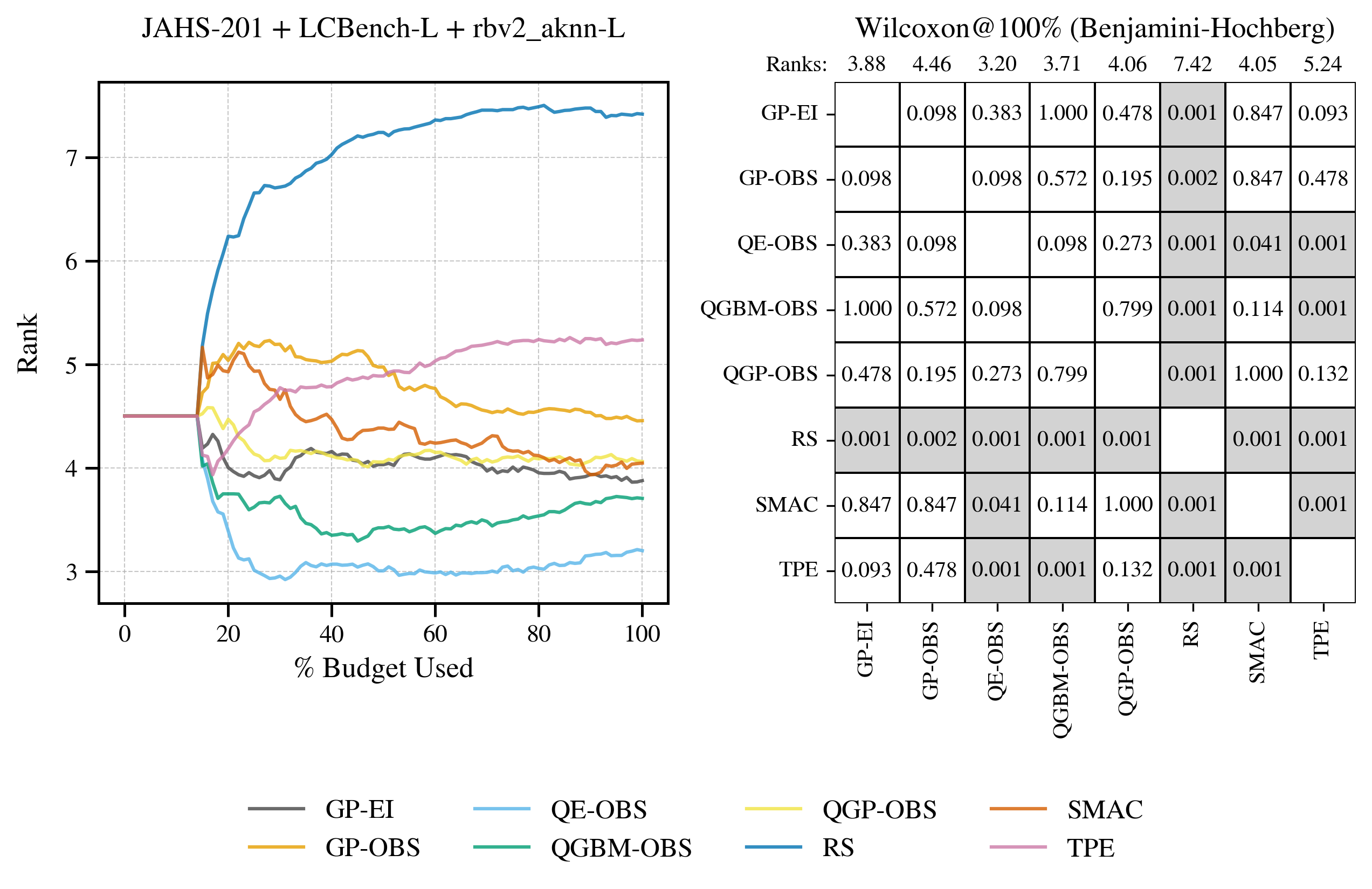}
    \caption{\textbf{Left:} search performance rank over iteration search budget for range of quantile and established HPO algorithms. Results cover 15 random warm start initializations. \textbf{Right:} Matrix of Wilcoxon Signed-Rank p-values per pairwise algorithm comparison at 100\% budget. P-values are adjusted for multiple comparison via Benhamini-Hochberg correction. Shaded cells denote significant comparisons.} 
    \label{sig-bench-general-iteration}
\end{figure}

Figure \ref{sig-bench-general-iteration} shows the performance of QE, QGP and QGBM alongside traditional ARD Gaussian Processes (GP), Tree Parzen Estimators (TPE) and SMAC across \textit{JAHS-Bench-201}, \textit{LCBench-L} and \textit{rbv2\_aknn-L}.

Quantile methods perform extremely competitively, with QE and QGBM placing in first and second place respectively. QGP ties with Expected Improvement based GP, and meaningfully outperforms its OBS GP equivalent (though the gap narrows by the end of the budget). SMAC is not distantly behind QGP, while TPE performs most poorly. 

Wilcoxon Signed-Rank significance analysis shows QE-OBS achieving significantly higher performance than SMAC and TPE, and near significant outperformance over QGBM-OBS and GP-OBS. QGBM-OBS significantly outperforms TPE, and near significantly outperforms SMAC. All other methods only significantly outperform random search.

\begin{figure}[t]
    \centering
    \includegraphics[scale=0.5]{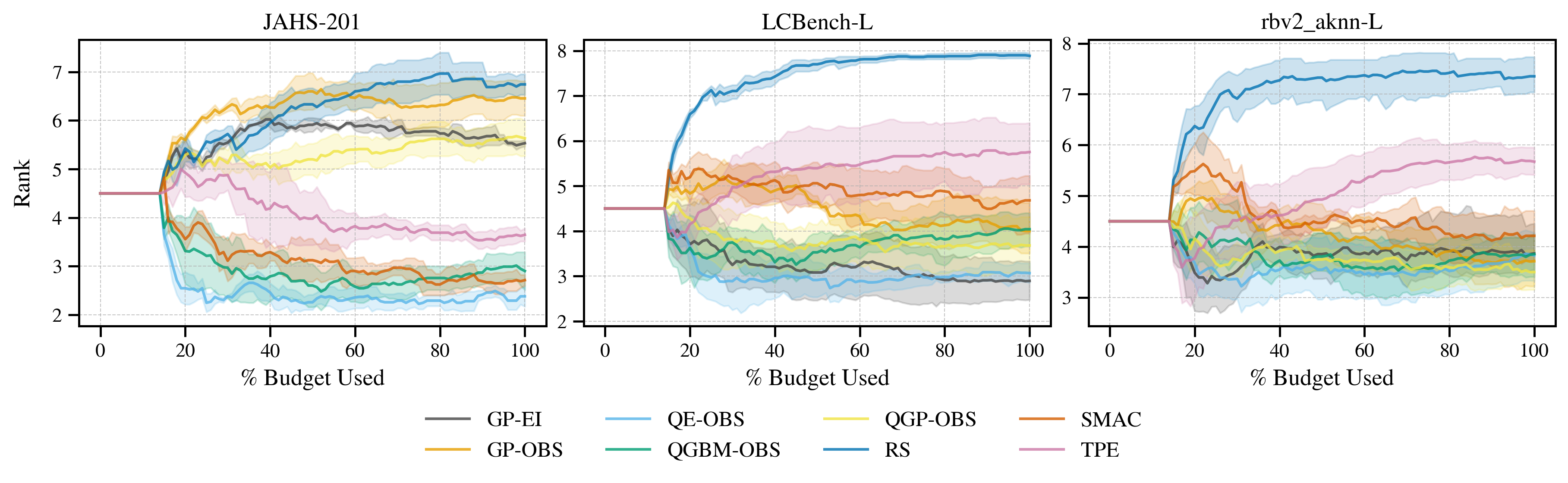}
    \caption{Search performance rank over iteration search budget for range of quantile and established HPO algorithms, segmented by benchmarking environment (columns). Results cover 15 random warm start initializations. Shaded region represents 95\% dataset-bootstrapped interval.} 
    \label{bench-general-individual-iteration}
\end{figure}

Beyond aggregate performance, Figure \ref{bench-general-individual-iteration} shows search performance breakdowns by individual benchmark, with important variations across environments. GP based methods, whether distributional or quantile-based, struggle significantly on the highly categorical \textit{JAHS-Bench-201} benchmark, with performances only marginally superior to random search. In this environment, SMAC's tree estimation confers it a significant performance boost (with performance approaching that of QGBM), though QE still meaningfully outperforms all surrogates. Remaining benchmarks show strong GP-EI performance, whose low global rank is primarily due to drag from \textit{JAHS-Bench-201}. On mostly continuous hyperparameter environments, GP-EI consistently performs similarly to QGBM, and is only narrowly outperformed by QE.

\paragraph{Stratified Analysis} Previous analysis shows comparative performance on the basis of large dataset benchmarks. This provided a general overview and identified key drivers of tree-based success on categorical hyperparameter environments.

\begin{figure}[t]
    \centering
    \includegraphics[scale=0.5]{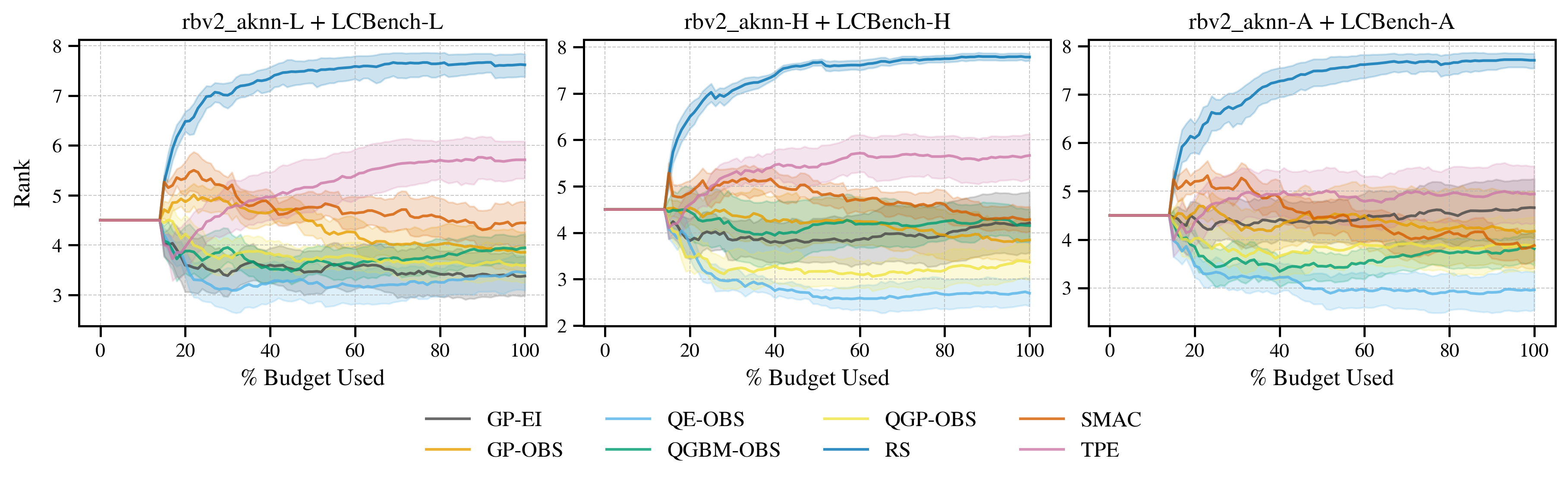}
    \caption{Search performance rank over iteration search budget for range of quantile and established HPO algorithms, segmented by benchmarking group (columns). Results cover 15 random warm start initializations. Shaded region represents 95\% dataset-bootstrapped interval.} 
    \label{bench-sub-iteration}
\end{figure}

To further stress the versatility of GPs, Figure \ref{bench-sub-iteration} compares search performance between previously explored large variants of the continuous \textit{LCBench} and \textit{rbv2\_aknn} environments and two variants of it that screen datasets for heteroskedasticity (-H) and asymmetry (-A).

The strong performance of GP-EI on the previously explored large benchmarks, is significantly weakened in heteroskedastic and asymmetric settings. QGP, on the other hand, is much more robust to these shifts, possibly due to the corrective effect of conformalization.

Interestingly, while QGBM is unaffected by shifting from large to asymmetric benchmarks (given quantile regression's ability to fit independent, non-distributional quantiles), it does deteriorate similarly to GP-EI in heteroskedastic settings. Lastly, QE continues to perform strongly, with first positions across all three benchmarks and a widening lead in both heteroskedastic and asymmetric settings.

This validates the usage of conformalized, quantile-based approaches in environments that challenge GP assumptions, as well as further validating the robustness and versatility of ensembles.

\section{Conclusion}
This study proposed enhancements to conformalized quantile hyperparameter optimization, while assessing the benefit of conformalization on both calibration and search performance. 

Acquisition function benchmarks revealed meaningful performance heterogeneity, with Thompson approaches outperforming Expected Improvement, and Optimistic Bayesian Sampling outperforming traditional Thompson Sampling.

Surrogate architecture comparisons highlighted strong benefits from ensembling, with QE outperforming alternatives. Conformalized Gaussian Processes and QGBM also performed competitively.

Selected combinations of quantile surrogates and acquisition functions were evaluated on a broader set of benchmarks against a range of popular HPO algorithms. Findings revealed meaningful, and frequently significant, outperformance by QE and QGBM architectures, with QE consistently achieving first or tied first place across all benchmarking environment groups. Performance gaps with GPs were shown to grow even larger on sub-populations of datasets with challenging categorical, heteroskedastic or asymmetric attributes.

Lastly, conformalization was shown to significantly improve local and marginal calibration quality on a greedy sampling simulation, with CV+ improving SCP performance, and DtACI improving ACI performance. These benefits were strongly transferable to search performance when sampling via Expected Improvement, but not when sampling via Thompson approaches.

\newpage

\begin{appendix}

\section{Calibration Performance by \textit{LCBench-L} Dataset}\label{Appendix - Calibration Performance by LCBench-L Dataset}

\begin{figure}[htbp]
    \centering
    \includegraphics[scale=0.45]{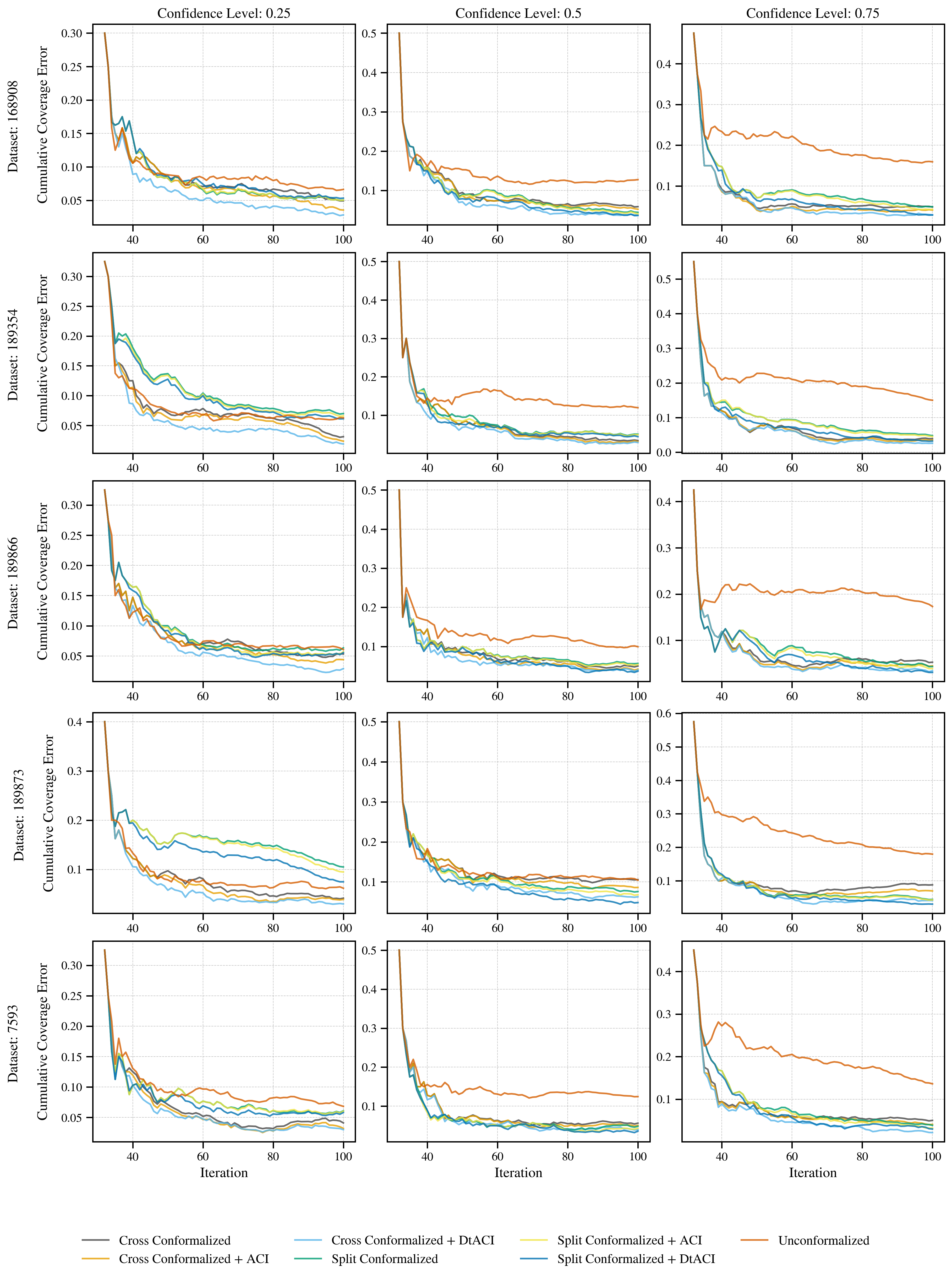}
    \caption{Cumulative coverage per search iteration across 25\%, 50\% and 75\% intervals from greedy expected value acquisition on \textit{LCBench-L} datasets. Results are averaged across 20 random warm started runs. Uncertainty regions mark 95\% dataset-bootstrapped intervals. Coverage reporting begins at iteration 32, post-conformalization.} 
    \label{multi-cov-cumulative}
\end{figure}

\newpage

\section{Runtime Aggregated Benchmarks}\label{Appendix - Runtime Aggregated Benchmarks}

Benchmark results in the main body of the paper are reported over a relativized iteration budget. Results in this appendix report the same results from the same simulations, but standardized over a relativized runtime budget.

\begin{figure}[htbp]
    \centering
    \includegraphics[scale=0.4]{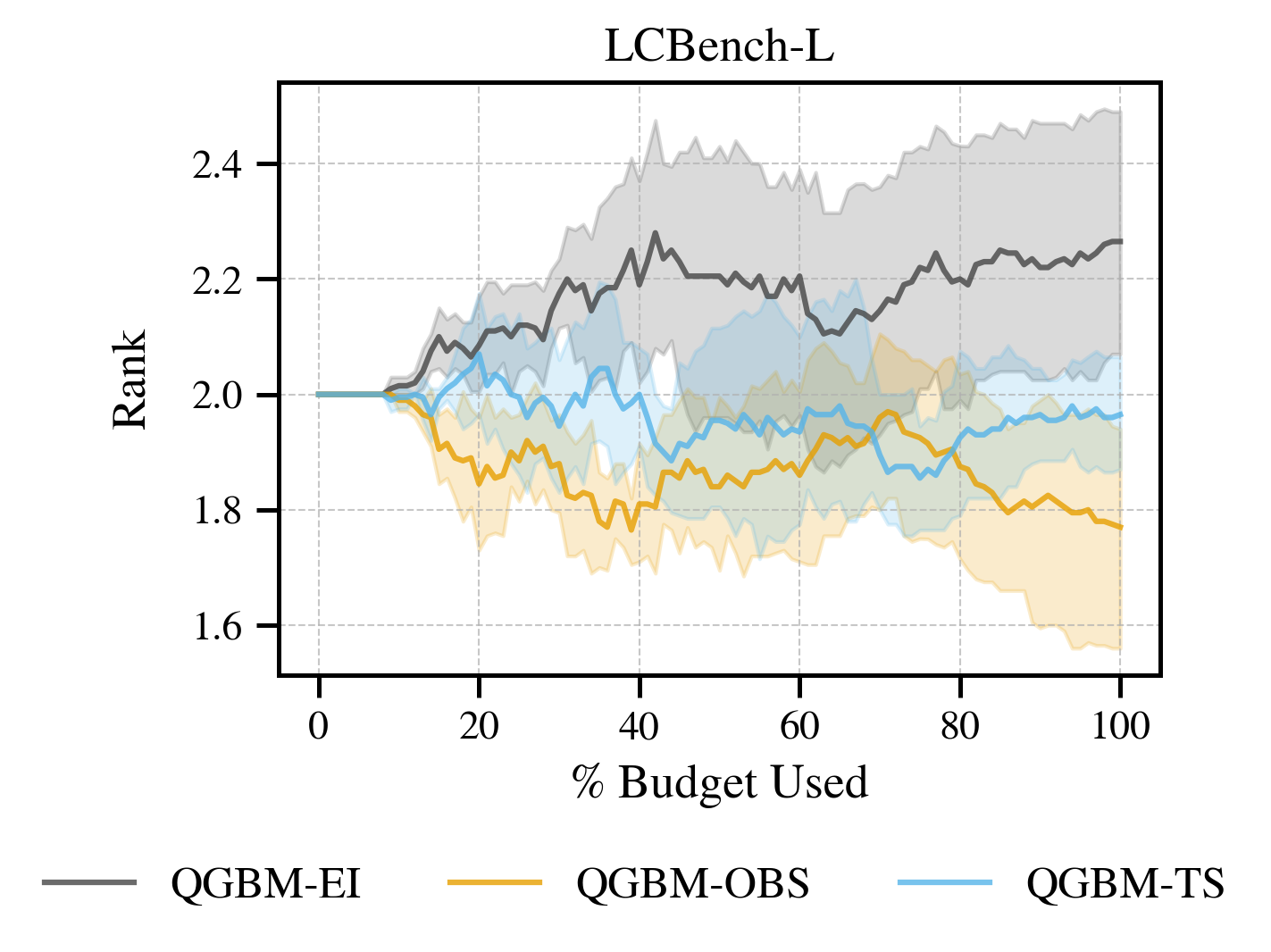}
    \caption{Search performance rank over runtime search budget per acquisition function on \textit{LCBench-L}, across 20 random warm start initializations. Shaded region represents 95\% dataset-bootstrapped interval.} 
    \label{af-runtime}
\end{figure}

\begin{figure}[htbp]
    \centering
    \includegraphics[scale=0.4]{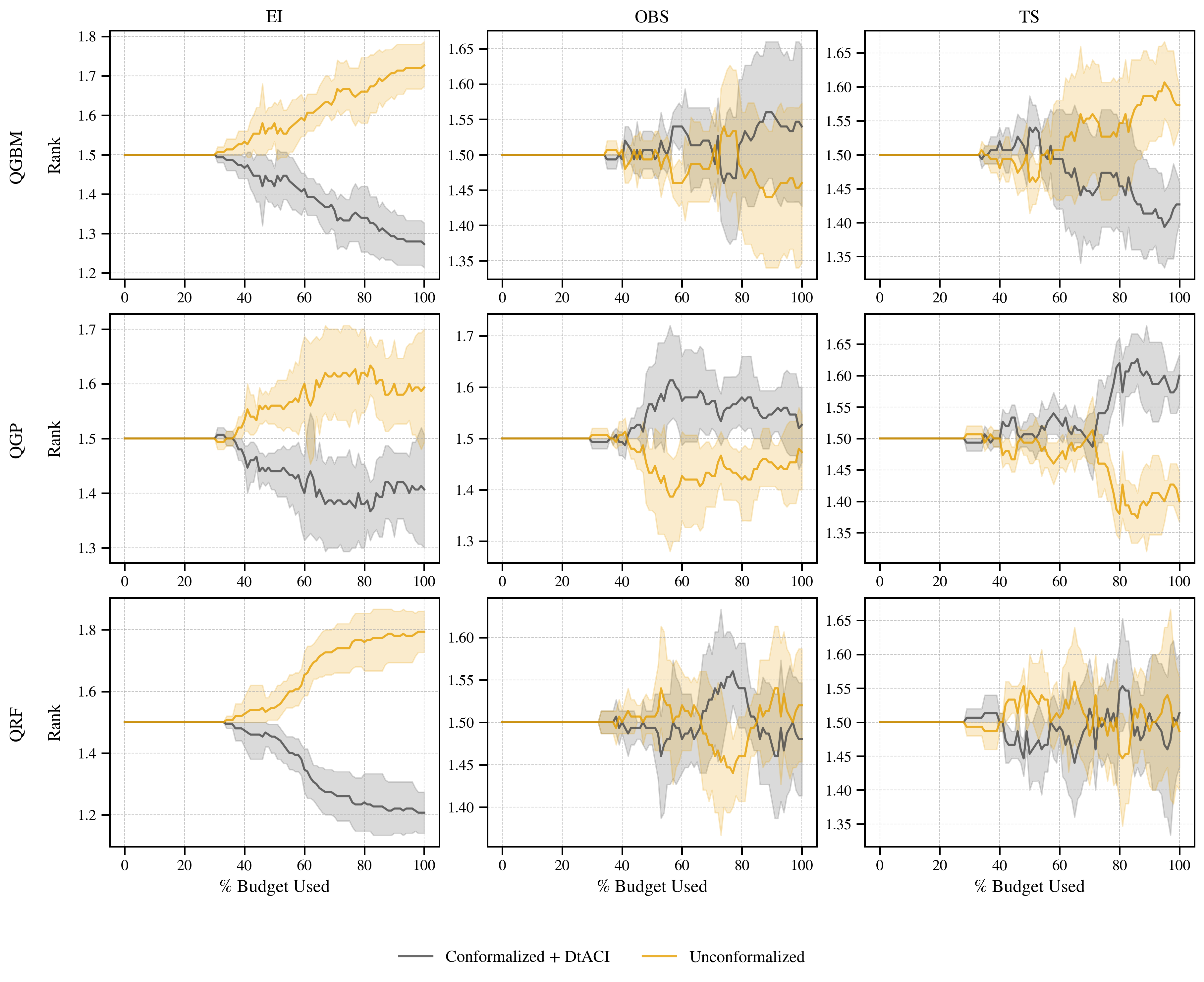}
    \caption{\textit{LCBench-L} search performance rank over runtime search budget. Performances are reported with and without conformalization, across several surrogate architectures (rows) and acquisition functions (columns). Results cover 20 random warm start initializations. Shaded region represents 95\% dataset-bootstrapped interval. Conformalization is carried out via CV+ up to the 50-th iteration, and SCP thereafter.} 
    \label{ce-runtime}
\end{figure}

\begin{figure}[htbp]
    \centering
    \includegraphics[scale=0.4]{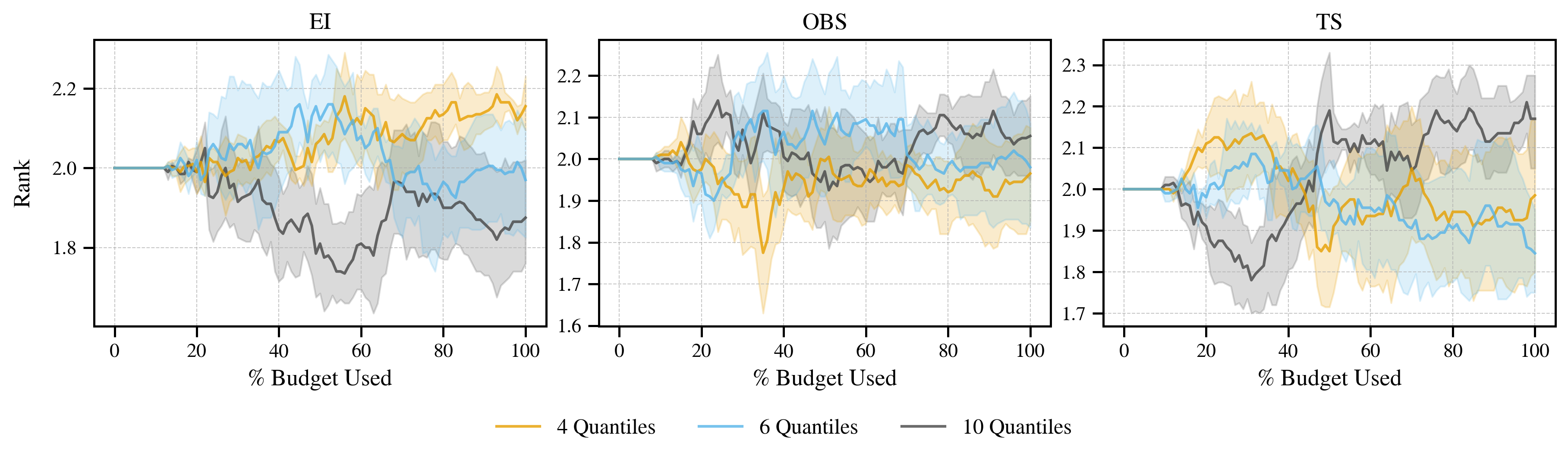}
    \caption{\textit{LCBench-L} search performance rank over runtime search budget for a QGBM surrogate, across multiple acquisition functions (columns) and quantile densities. Results cover 20 random warm start initializations. Shaded region represents 95\% dataset-bootstrapped interval.} 
    \label{nq-runtime}
\end{figure}

\begin{figure}[htbp]
    \centering
    \includegraphics[scale=0.4]{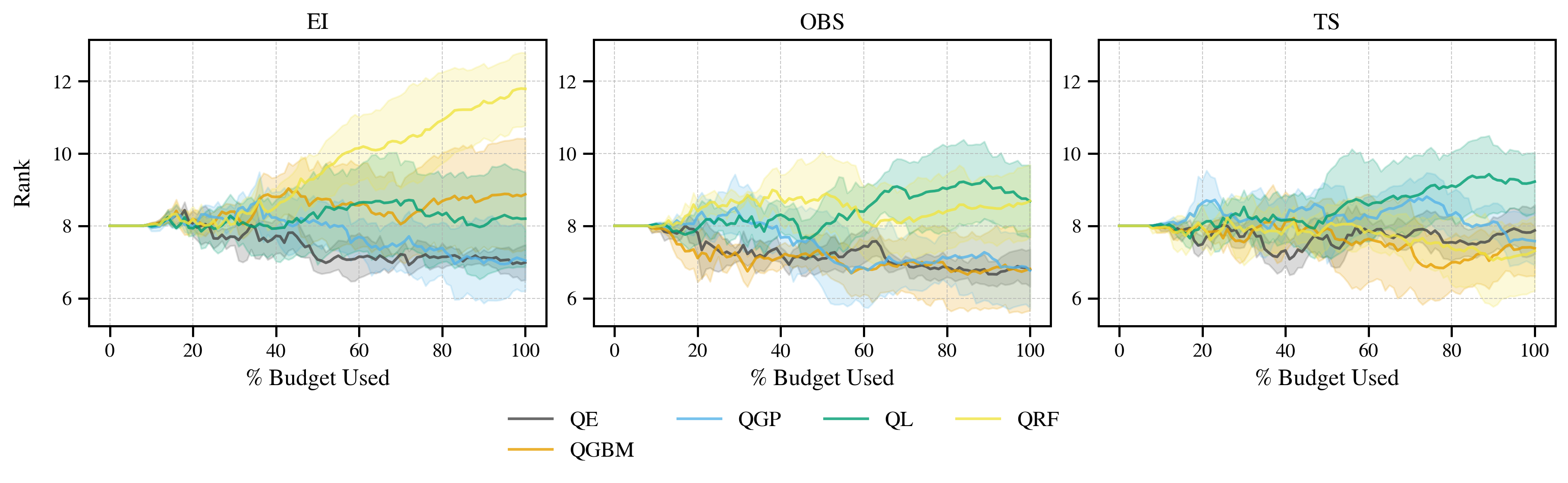}
    \caption{\textit{LCBench-L} search performance rank over runtime search budget for a range of surrogate architectures, across multiple acquisition functions (columns). Ranks are shared across plots (each surrogate and acquisition combination is treated as a ranking variant). Results cover 20 random warm start initializations. Shaded region represents 95\% dataset-bootstrapped interval.} 
    \label{dynamic-arch-runtime}
\end{figure}

\begin{figure}[htbp]
    \centering
    \includegraphics[scale=0.4]{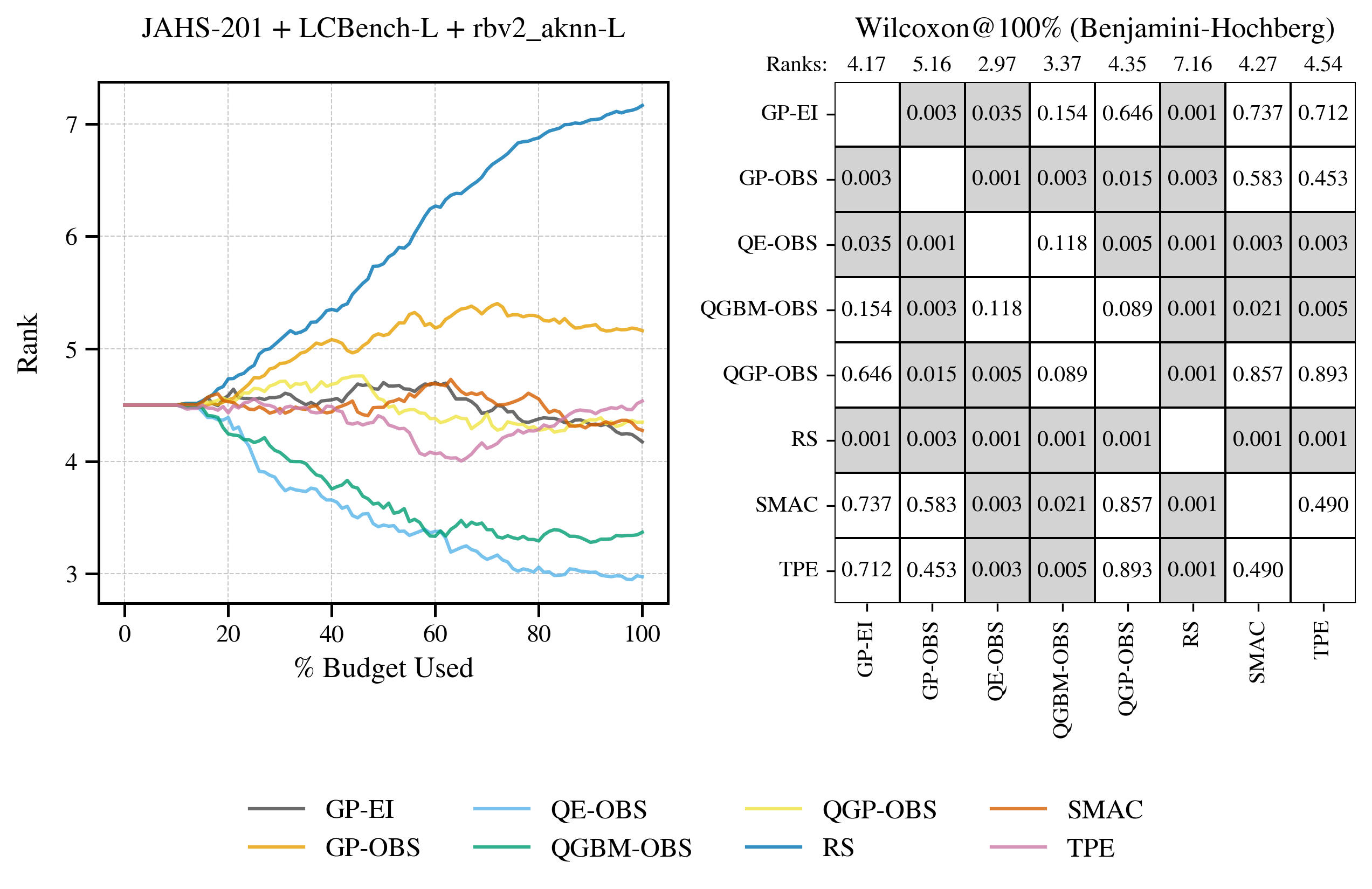}
    \caption{\textbf{Left:} search performance rank over runtime search budget for range of quantile and established HPO algorithms. Results cover 15 random warm start initializations. \textbf{Right:} Matrix of Wilcoxon Signed-Rank p-values per pairwise algorithm comparison at 100\% budget. P-values are adjusted for multiple comparison via Benhamini-Hochberg correction. Shaded cells denote significant comparisons.} 
    \label{sig-bench-general-runtime}
\end{figure}

\clearpage

\begin{figure}[H]
    \centering
    \includegraphics[scale=0.4]{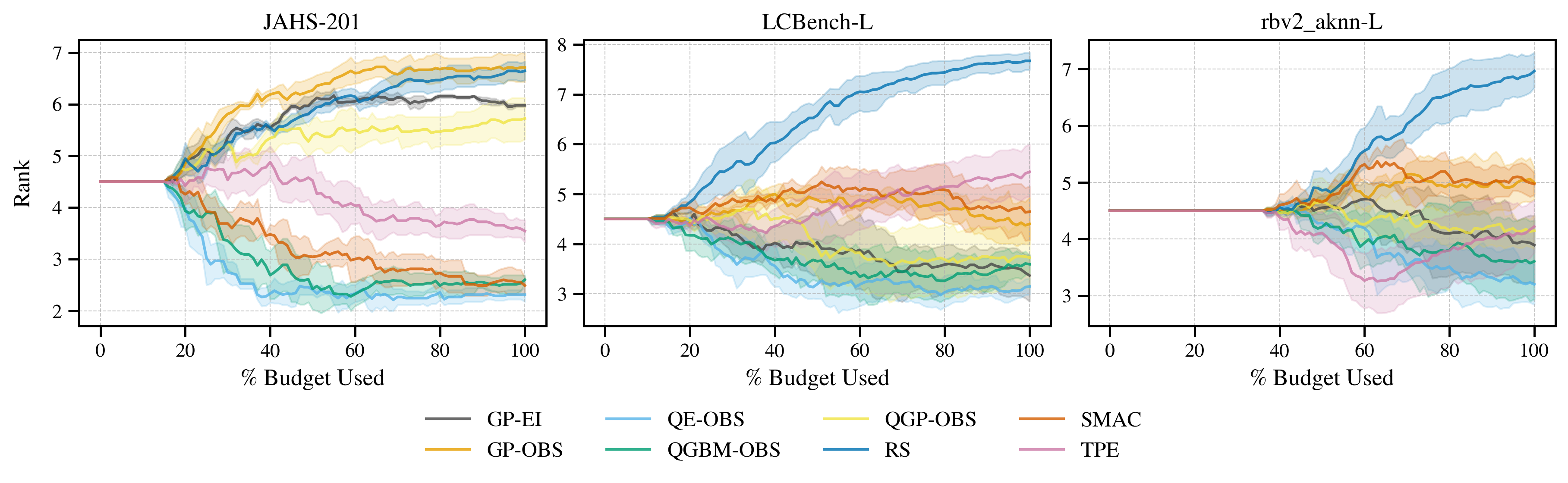}
    \caption{Search performance rank over runtime search budget for range of quantile and established HPO algorithms, segmented by benchmarking environment (columns). Results cover 15 random warm start initializations. Shaded region represents 95\% dataset-bootstrapped interval.} 
    \label{bench-general-individual-runtime}
\end{figure}

\begin{figure}[H]
    \centering
    \includegraphics[scale=0.4]{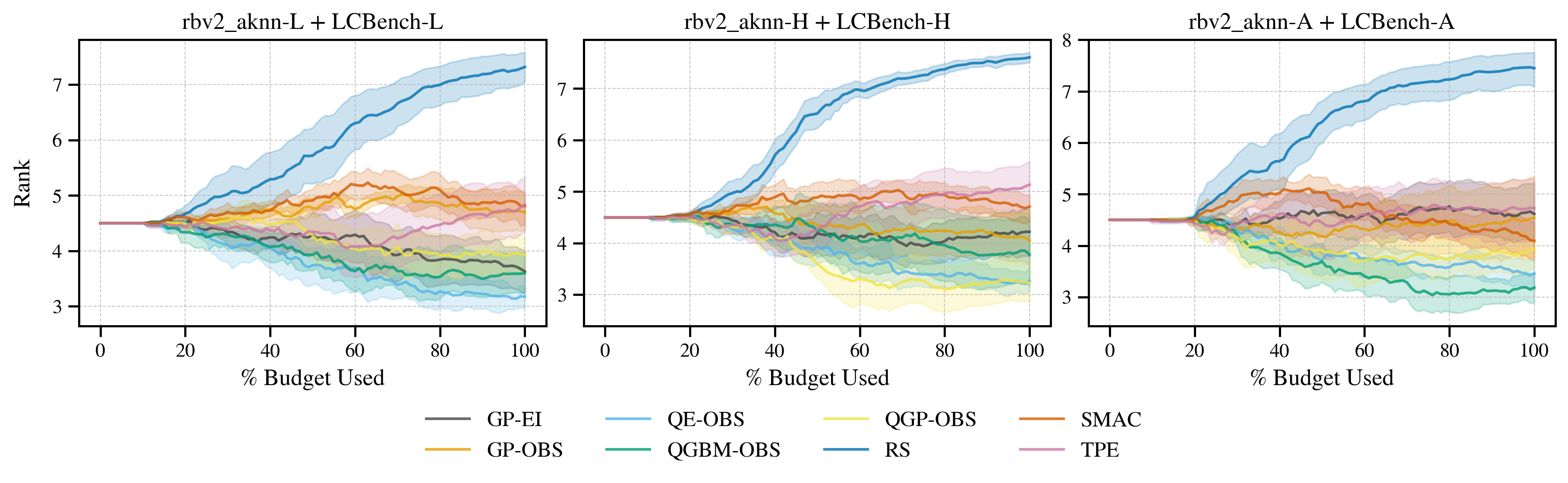}
    \caption{Search performance rank over runtime search budget for range of quantile and established HPO algorithms, segmented by benchmarking group (columns). Results cover 15 random warm start initializations. Shaded region represents 95\% dataset-bootstrapped interval.} 
    \label{bench-sub-runtime}
\end{figure}

\clearpage

\section{OpenML Stratifications}\label{Appendix - OpenML Stratifications}
OpenML identifiers for each dataset constituent of each benchmark stratification can be found below:
\begin{itemize}
    \item \textbf{LCBench-L:} 189873, 168908, 7593, 189866, 189354.
    \item \textbf{LCBench-H:} 168331, 189866, 167181, 126026, 7593.
    \item \textbf{LCBench-A:} 189873, 167185, 167152, 146212, 168910.
    \item \textbf{rbv2\_aknn-L:} 40927, 41162, 40923, 41165, 41161.
    \item \textbf{rbv2\_aknn-H:} 41138, 1478, 554, 1486, 41027.
    \item \textbf{rbv2\_aknn-A:} 40978, 1461, 300, 1040, 41157.
\end{itemize}

\section{Code}\label{Appendix - Code}

Benchmarking code to reproduce results from this paper are stored in the \textit{arxiv-ecqr-2025-v1} branch of the following GitHub repository: \href{https://github.com/rick12000/hpo-benchmark}{https://github.com/rick12000/hpo-benchmark}.

As stated in the benchmarking repository's \textit{README.md}, all conformalized quantile HPO algorithms proposed in this repository can be accessed from the \textit{ConfOpt} package, either via \href{https://pypi.org/project/confopt}{PyPI} or at the following GitHub repository: \href{https://github.com/rick12000/confopt}{https://github.com/rick12000/confopt}.

\section{Abbreviations}\label{Appendix - Abbreviations}
\begin{itemize}
    \item HPO: Hyperparameter Optimization
    \item TPE: Tree-Parzen Estimator
    \item GP: Gaussian Process
    \item EI: Expected Improvement
    \item TS: Thompson Sampling
    \item OBS: Optimistic Bayesian Sampling
    \item SCP: Split Conformal Prediction
    \item CQR: Conformalized Quantile Regression
    \item QGBM: Quantile Gradient Boosted Machines
    \item QRF: Quantile Regression Forest
    \item QGP: Quantile Gaussian Process
    \item SOTA: State of the Art
\end{itemize}

\end{appendix}
\newpage

\bibliographystyle{unsrt}
\bibliography{main} 

\begin{thebibliography}{10}

\bibitem{NIPS2012_05311655}
Jasper Snoek, Hugo Larochelle, and Ryan~P Adams.
\newblock Practical bayesian optimization of machine learning algorithms.
\newblock In F.~Pereira, C.J. Burges, L.~Bottou, and K.Q. Weinberger, editors, {\em Advances in Neural Information Processing Systems}, volume~25. Curran Associates, Inc., 2012.

\bibitem{Rasmussen2004}
Carl~Edward Rasmussen.
\newblock {\em Gaussian Processes in Machine Learning}, pages 63--71.
\newblock Springer Berlin Heidelberg, Berlin, Heidelberg, 2004.

\bibitem{movckus1975bayesian}
Jonas Mo{\v{c}}kus.
\newblock On {B}ayesian methods for seeking the extremum.
\newblock In {\em Optimization Techniques IFIP Technical Conference: Novosibirsk, July 1--7, 1974}, pages 400--404. Springer, 1975.

\bibitem{JMLR:v13:bergstra12a}
James Bergstra and Yoshua Bengio.
\newblock Random search for hyper-parameter optimization.
\newblock {\em Journal of Machine Learning Research}, 13(10):281--305, 2012.

\bibitem{NIPS2011_86e8f7ab}
James Bergstra, R\'{e}mi Bardenet, Yoshua Bengio, and Bal\'{a}zs K\'{e}gl.
\newblock Algorithms for hyper-parameter optimization.
\newblock In J.~Shawe-Taylor, R.~Zemel, P.~Bartlett, F.~Pereira, and K.Q. Weinberger, editors, {\em Advances in Neural Information Processing Systems}, volume~24. Curran Associates, Inc., 2011.

\bibitem{10.1007/978-3-642-25566-3_40}
Frank Hutter, Holger~H. Hoos, and Kevin Leyton-Brown.
\newblock Sequential model-based optimization for general algorithm configuration.
\newblock In Carlos A.~Coello Coello, editor, {\em Learning and Intelligent Optimization}, pages 507--523, Berlin, Heidelberg, 2011. Springer Berlin Heidelberg.

\bibitem{breiman2014random}
Leo Breiman and Adele Cutler.
\newblock Random forests. 2001.
\newblock {\em Mach. Learn}, 45(5), 2014.

\bibitem{10.2307/1913643}
Roger Koenker and Gilbert Bassett.
\newblock Regression quantiles.
\newblock {\em Econometrica}, 46(1):33--50, 1978.

\bibitem{a05bb90f8fb8491392d9564461103e48}
\{Jeroen van\} Hoof and Joaquin Vanschoren.
\newblock Hyperboost: Hyperparameter optimization by gradient boosting surrogate models.
\newblock {\em CoRR}, abs/2101.02289, 2021.

\bibitem{pmlr-v119-salinas20a}
David Salinas, Huibin Shen, and Valerio Perrone.
\newblock A quantile-based approach for hyperparameter transfer learning.
\newblock In Hal~Daumé III and Aarti Singh, editors, {\em Proceedings of the 37th International Conference on Machine Learning}, volume 119 of {\em Proceedings of Machine Learning Research}, pages 8438--8448. PMLR, 13--18 Jul 2020.

\bibitem{JMLR:v9:shafer08a}
Glenn Shafer and Vladimir Vovk.
\newblock A tutorial on conformal prediction.
\newblock {\em Journal of Machine Learning Research}, 9(12):371--421, 2008.

\bibitem{doyle2023achoadaptiveconformalhyperparameter}
Riccardo Doyle.
\newblock Acho: Adaptive conformal hyperparameter optimization, 2023.

\bibitem{Lei03072018}
Jing Lei, Max G’Sell, Alessandro Rinaldo, Ryan~J. Tibshirani, and Larry Wasserman.
\newblock Distribution-free predictive inference for regression.
\newblock {\em Journal of the American Statistical Association}, 113(523):1094--1111, 2018.

\bibitem{NEURIPS2019_5103c358}
Yaniv Romano, Evan Patterson, and Emmanuel Candes.
\newblock Conformalized quantile regression.
\newblock In H.~Wallach, H.~Larochelle, A.~Beygelzimer, F.~d\textquotesingle Alch\'{e}-Buc, E.~Fox, and R.~Garnett, editors, {\em Advances in Neural Information Processing Systems}, volume~32. Curran Associates, Inc., 2019.

\bibitem{salinas2023optimizing}
D.~Salinas, J.~Golebiowski, A.~Klein, M.~Seeger, and C.~Archambeau.
\newblock Optimizing hyperparameters with conformal quantile regression.
\newblock In {\em Proceedings of the 40th International Conference on Machine Learning}, ICML'23. JMLR.org, 2023.

\bibitem{4a848dd1-54e3-3c3c-83c3-04977ded2e71}
Jerome~H. Friedman.
\newblock Greedy function approximation: A gradient boosting machine.
\newblock {\em The Annals of Statistics}, 29(5):1189--1232, 2001.

\bibitem{dc35850b-2ca1-314f-9e0d-470713436b17}
William~R. Thompson.
\newblock On the likelihood that one unknown probability exceeds another in view of the evidence of two samples.
\newblock {\em Biometrika}, 25(3/4):285--294, 1933.

\bibitem{journals/ml/AuerCF02}
Peter Auer, Nicolò Cesa-Bianchi, and Paul Fischer.
\newblock Finite-time analysis of the multiarmed bandit problem.
\newblock {\em Mach. Learn.}, 47(2-3):235--256, 2002.

\bibitem{JMLR:v13:may12a}
Benedict~C. May, Nathan Korda, Anthony Lee, and David~S. Leslie.
\newblock Optimistic bayesian sampling in contextual-bandit problems.
\newblock {\em Journal of Machine Learning Research}, 13(67):2069--2106, 2012.

\bibitem{gibbs2021adaptive}
Isaac Gibbs and Emmanuel Candes.
\newblock Adaptive conformal inference under distribution shift.
\newblock In A.~Beygelzimer, Y.~Dauphin, P.~Liang, and J.~Wortman Vaughan, editors, {\em Advances in Neural Information Processing Systems}, 2021.

\bibitem{JMLR:v25:22-1218}
Isaac Gibbs and Emmanuel~J. Cand{{\`e}}s.
\newblock Conformal inference for online prediction with arbitrary distribution shifts.
\newblock {\em Journal of Machine Learning Research}, 25(162):1--36, 2024.

\bibitem{51791361-8fe2-38d5-959f-ae8d048b490d}
Robert Tibshirani.
\newblock Regression shrinkage and selection via the lasso.
\newblock {\em Journal of the Royal Statistical Society. Series B (Methodological)}, 58(1):267--288, 1996.

\bibitem{JMLR:v7:meinshausen06a}
Nicolai Meinshausen.
\newblock Quantile regression forests.
\newblock {\em Journal of Machine Learning Research}, 7(35):983--999, 2006.

\bibitem{WOLPERT1992241}
David~H. Wolpert.
\newblock Stacked generalization.
\newblock {\em Neural Networks}, 5(2):241--259, 1992.

\bibitem{barber2020predictiveinferencejackknife}
Rina~Foygel Barber, Emmanuel~J. Candes, Aaditya Ramdas, and Ryan~J. Tibshirani.
\newblock Predictive inference with the jackknife+, 2020.

\bibitem{NEURIPS2022_fd78f2f6}
Archit Bansal, Danny Stoll, Maciej Janowski, Arber Zela, and Frank Hutter.
\newblock Jahs-bench-201: A foundation for research on joint architecture and hyperparameter search.
\newblock In S.~Koyejo, S.~Mohamed, A.~Agarwal, D.~Belgrave, K.~Cho, and A.~Oh, editors, {\em Advances in Neural Information Processing Systems}, volume~35, pages 38788--38802. Curran Associates, Inc., 2022.

\bibitem{krizhevsky2009learning}
Alex Krizhevsky and Geoffrey Hinton.
\newblock Learning multiple layers of features from tiny images.
\newblock Technical Report~0, University of Toronto, Toronto, Ontario, 2009.

\bibitem{kather2016multi}
Jakob~Nikolas Kather, Cleo-Aron Weis, Francesco Bianconi, Susanne~M Melchers, Lothar~R Schad, Timo Gaiser, Alexander Marx, and Frank~Gerrit Z{"o}llner.
\newblock Multi-class texture analysis in colorectal cancer histology.
\newblock {\em Scientific reports}, 6:27988, 2016.

\bibitem{Xiao2017FashionMNISTAN}
Han Xiao, Kashif Rasul, and Roland Vollgraf.
\newblock Fashion-mnist: a novel image dataset for benchmarking machine learning algorithms.
\newblock {\em ArXiv}, abs/1708.07747, 2017.

\bibitem{ZimLin2021a}
Lucas Zimmer, Marius Lindauer, and Frank Hutter.
\newblock Auto-pytorch tabular: Multi-fidelity metalearning for efficient and robust autodl.
\newblock {\em IEEE Transactions on Pattern Analysis and Machine Intelligence}, 43(9):3079 -- 3090, 2021.

\bibitem{pmlr-v188-pfisterer22a}
Florian Pfisterer, Lennart Schneider, Julia Moosbauer, Martin Binder, and Bernd Bischl.
\newblock Yahpo gym - an efficient multi-objective multi-fidelity benchmark for hyperparameter optimization.
\newblock In Isabelle Guyon, Marius Lindauer, Mihaela van~der Schaar, Frank Hutter, and Roman Garnett, editors, {\em Proceedings of the First International Conference on Automated Machine Learning}, volume 188 of {\em Proceedings of Machine Learning Research}, pages 3/1--39. PMLR, 25--27 Jul 2022.

\bibitem{10.1109/TPAMI.2018.2889473}
Yu~A. Malkov and D.~A. Yashunin.
\newblock Efficient and robust approximate nearest neighbor search using hierarchical navigable small world graphs.
\newblock {\em IEEE Trans. Pattern Anal. Mach. Intell.}, 42(4):824–836, April 2020.

\bibitem{1053964}
T.~Cover and P.~Hart.
\newblock Nearest neighbor pattern classification.
\newblock {\em IEEE Transactions on Information Theory}, 13(1):21--27, 1967.

\end{thebibliography}

\end{document}